%% file: ms.tex
\documentclass[]{./include/latex-class-interact/interact}

\usepackage[utf8]{inputenc}
\usepackage{amsmath}
\usepackage{amssymb}
\usepackage{microtype}
\usepackage[all]{xy}
\usepackage[natbibapa,nodoi]{apacite}
\usepackage{url}
\usepackage{hyperref}
\usepackage{enumitem}

\input{./include/asp-macros/ht}
\input{./include/asp-macros/dtel}
\input{./include/asp-macros/logics}

\input{./include/asp-macros/systems}

\input{macro}
\newtheorem{definition}{Definition}

\newtheorem{proposition}{Proposition}
\newtheorem{corollary}{Corollary}

\begin{document}

\title{Metric Dynamic Equilibrium Logic}

\author{%
  \name{Arvid Becker\textsuperscript{a}\thanks{CONTACT Arvid Becker. Email: {arvid.becker@uni-potsdam.de}},
    Pedro Cabalar\textsuperscript{b},
    Mart{\'{i}}n Di{\'e}guez\textsuperscript{c},
    Luis Fari{\~n}as\textsuperscript{d},
    Torsten Schaub\textsuperscript{a},
    Anna Schuhmann\textsuperscript{a}}
  \affil{\textsuperscript{a}University of Potsdam, Germany;
    \textsuperscript{b}University of Corunna, Spain;
    \textsuperscript{c}Universit\'e d'Angers, France;
	\textsuperscript{d}Universit\'e de Toulouse, France
	}
}

\maketitle

\input{./abstract}
\input{./introduction}
\input{./mdht}
\input{./mdel}
\input{./relation}
\input{./discussion}

\input{./acknowledgments}
\bibliographystyle{apacite}
\bibliography{./include/bibliography/krr,./local,./include/bibliography/procs}
\appendix
\input{./proofs}

\end{document}

%% file: include/asp-macros/ht.tex
\newcommand{\tuple}[1]{\ensuremath{\langle #1 \rangle}}
\newcommand{\Htrace}{\ensuremath{\mathbf{H}}}
\newcommand{\Ttrace}{\ensuremath{\mathbf{T}}}
\newcommand{\M}{\ensuremath{\mathbf{M}}}


%% file: include/asp-macros/logics.tex
\providecommand{\logfont}{\textrm}

\newcommand{\HT}{\ensuremath{\logfont{HT}}}

\newcommand{\LTL}{\ensuremath{\logfont{LTL}}}

\newcommand{\THT}{\ensuremath{\logfont{THT}}}

\newcommand{\TEL}{\ensuremath{\logfont{TEL}}}

\newcommand{\LDL}{\ensuremath{\logfont{LDL}}}

\newcommand{\DHT}{\ensuremath{\logfont{DHT}}}

\newcommand{\DEL}{\ensuremath{\logfont{DEL}}}

\newcommand{\DL}{\ensuremath{\logfont{DL}}}

\newcommand{\MTL}{\ensuremath{\logfont{MTL}}}

\newcommand{\MHT}{\ensuremath{\logfont{MHT}}}

\newcommand{\MEL}{\ensuremath{\logfont{MEL}}}

%% file: include/asp-macros/systems.tex
\providecommand{\sysfont}{\textit}

\newcommand{\clingo}{\sysfont{clingo}}

\newcommand{\telingo}{\sysfont{telingo}}

%% file: macro.tex
\newcommand{\eqdef}{\ensuremath{\mathbin{\raisebox{-1pt}[-3pt][0pt]{$\stackrel{\mathit{def}}{=}$}}}}
\newtheorem{remark}{Remark}
\newcommand{\myatom}{\ensuremath{p}}
\newcommand{\PV}{\ensuremath{\mathcal{A}}}

\newcommand{\EM}{\ensuremath{\mathrm{EM}}}

\newcommand{\mdalways}[2]{\ensuremath{[#1]_{#2}\,}}
\newcommand{\mdeventually}[2]{\ensuremath{\langle#1\rangle_{#2}\,}}

\newcommand{\mdalwaysF}[3]{\ensuremath{[#1]_{#2}\,{#3}}}                     
\newcommand{\mdeventuallyF}[3]{\ensuremath{\langle#1\rangle_{#2}\,{#3}}}     

\newcommand{\mdrel}[2]{\ensuremath{{\parallel}{#1}{\parallel}^{#2}}}

\newcommand{\tmf}{\ensuremath{\tau}}

\newcommand{\MDL}{\ensuremath{\mathrm{MDL}}}

\newcommand{\MDHT}{\ensuremath{\mathrm{MDHT}}}
\newcommand{\MDHTf}{\ensuremath{\mathrm{MDHT}_{\!f}}}
\newcommand{\MDHTo}{\ensuremath{\mathrm{MDHT}_{\!\omega}}}
\newcommand{\MDEL}{\ensuremath{\mathrm{MDEL}}}
\newcommand{\MDELf}{\ensuremath{{\MDEL}_{\!f}}}
\newcommand{\MDELo}{\ensuremath{{\MDEL}_{\omega}}}

\newcommand{\metricI}[1]{\ensuremath{{#1}_{I}}}
\newcommand{\metric}[3]{\ensuremath{{#1}_{
    \ifthenelse{\equal{#2}{#3}}         
    {#2}                                
    {                                   
      \ifthenelse{\equal{#2}{0}}        
      {                                 
        \ifthenelse{\equal{#3}{\omega}} 
        {}                              
        {\leq#3}                        
      }
      {                                 
        \ifthenelse{\equal{#3}{\omega}} 
        {\geq#2}                        
        {\intervoo{#2}{#3}}              
      }
    }}}}

\newcommand{\Proposition}[1]{\textbf{(\textit{Proposition~\ref{#1}})}\par}
\newcommand{\Corollary}[1]{\textbf{(\textit{Corollary~\ref{#1}})}\par}

\newcommand{\invert}[1]{\ensuremath{{#1}^-}}
\newcommand{\converse}[1]{\ensuremath{{#1}^-}}
\newcommand{\kstar}[1]{\ensuremath{{#1}^\ast}}

\newtheorem{example}{Example}


%% file: abstract.tex
\begin{abstract}
  In temporal extensions of Answer Set Programming (ASP) based on linear-time,
  the behavior of dynamic systems is captured by sequences of states.
  While this representation reflects their relative order,
  it abstracts away the specific times associated with each state.
  In many applications, however, timing constraints are important like, for instance, when planning and scheduling go hand in hand.
  We address this by developing a metric extension of linear-time Dynamic Equilibrium Logic,
  in which dynamic operators are constrained by intervals over integers.
  The resulting Metric Dynamic Equilibrium Logic provides the foundation of an ASP-based approach for specifying
  qualitative and quantitative dynamic constraints.
  As such, it constitutes the most general among a whole spectrum of temporal extensions of Equilibrium Logic.
  In detail, we show that it encompasses Temporal, Dynamic, Metric, and regular Equilibrium Logic,
  as well as its classic counterparts once the law of the excluded middle is added.
\end{abstract}
%

%% file: introduction.tex
\section{Introduction}\label{sec:introduction}

Reasoning about action and change, or more generally reasoning about dynamic systems,
is not only central to knowledge representation and reasoning
but at the heart of computer science~\citep{TIMEHandbook}.
In practice, this often requires both qualitative as well as quantitative dynamic constraints.
For instance, when planning and scheduling go hand in hand,
actions may have durations and their effects may need to meet deadlines.

Over the last years, we addressed qualitative dynamic constraints by combining traditional approaches,
like Dynamic and Linear Temporal Logic (\DL~\citep{hatiko00a} and \LTL~\citep{pnueli77a}),
with the base logic of Answer Set Programming (ASP~\citep{lifschitz99b}), namely,
the logic of Here-and-There (\HT~\citep{heyting30a}) and its non-monotonic extension, called Equilibrium Logic~\citep{pearce96a}. 
This resulted in non-monotonic linear dynamic and temporal equilibrium logics
(\DEL~\citep{bocadisc18a,cadisc19a} and \TEL~\citep{cabper07a,agcadipevi13a,cakascsc18a,agcadipescscvi20a})
that gave rise to the temporal ASP system \telingo~\citep{cakamosc19a,cadilasc20a} extending the ASP system
\clingo~\citep{gekakaosscwa16a}.

A commonality of dynamic and temporal logics is that they abstract from specific time points when capturing temporal relationships.
For instance, in temporal logic, we can use the formula
\(
\alwaysF ( \mathit{use} \to \eventuallyF \mathit{clean})
\)
to express that at any time a machine has to be eventually cleaned after being used.
Nothing can be said about the delay between using and cleaning the machine.

A key design decision was to base both logics, \TEL\ and \DEL, on the same linear-time semantics
embodied by sequences of states.
%
We continued to maintain the same linear-time semantics
when elaborating upon a first ``light-weight'' metric temporal extension of \HT~\citep{cadiscsc20a}.
The ``light-weightiness'' is due to treating time as a state counter by identifying the next time with the next state.
For instance, this allows us to refine our example by expressing that, if the machine is used,
it has to be cleaned within the next 3 states, viz.\
\(
\alwaysF (\mathit{use} \to \metric{\eventuallyF}{1}{3}\mathit{clean})
\).
Although this permits the restriction of temporal operators to subsequences of states,
no fine-grained timing constraints are expressible.

In \citep{cadiscsc22a},
we filled this gap in the context of temporal logic
by associating each state with its \emph{time}, as done in Metric Temporal Logic (\MTL~\citep{koymans90a}).
This resulted in a non-monotonic metric temporal extension of \HT, referred to as \MEL.
It allows us to measure time differences between events.
For instance, in our example, we may thus express that whenever the machine is used, it has to be cleaned within 60 to 120 time units, by writing:
\begin{align}\label{ex:use:clean}
  \alwaysF (\mathit{use} \to \metric{\eventuallyF}{60}{120}\mathit{clean}) \ .
\end{align}
Unlike the non-metric version,
this stipulates that
once $\mathit{use}$ is true in a state,
$\mathit{clean}$ must be true in some future state
whose associated time is at least 60 and at most 120 time units after the time of $\mathit{use}$.
The choice of time domain is crucial, and might even lead to undecidability in the continuous case (that is, using real numbers).
We rather adapt a discrete approach that offers a sequence of snapshots of a dynamic system.

In this paper,
we combine the aforementioned temporal, dynamic, and (time-based) metric extensions of the logic of Here-and-There and its
non-monotonic extension Equilibrium logic
within a single logical setting by extending the dynamic variants with time-based metrics.
This results in the Metric Dynamic logic of Here-and-There (\MDHT) and its non-monotonic extension of Metric Dynamic Equilibrium Logic (\MDEL).
In this setting, we can express the statement in \eqref{ex:use:clean} as
\begin{align*}
  \mdalwaysF{\kstar{\stp}}{\intervoo{-\omega}{\omega}}{\left(\mdalwaysF{\mathit{use}?}{\intervoo{-\omega}{\omega}}{\mdeventuallyF{\kstar{\stp}}{\intervoo{60}{120}}{\mathit{clean}}}\right)} \ .
\end{align*}
As detailed below,
we may thus formulate temporal, metric, as well as Boolean operators in terms of metric dynamic ones.
This already hints at the great expressive power of \MDHT\ and \MDEL.

The main contribution of the paper is the definition of \MDHT\ itself.
Its syntax and semantics were carefully designed to cover \emph{both} the dynamic operators and the metric intervals, that constituted different and independent extensions, in a way that their combination could also provide other meaningful representational choices.
In this way, for instance, we propose intervals that may use negative positions (something uncommon in the literature) not only to accommodate past time operators in a more natural way, but also to offer more flexibility to treat relative time windows for a given expression.
The main results prove that the previous temporal approaches, \TEL, \DEL\ and \MEL, are now fragments of the more general formalism \MDEL.

The rest of the paper is organized as follows.
We start by defining the syntax, semantics, and some properties of \MDHT\ in Section~\ref{sec:mdht}
and define \MDEL\ in Section~\ref{sec:mdel}.
We then show in Section~\ref{sec:relation} how an existing (monotonic) metric dynamic logic~\citep{bakrtr17a} as well as
the various aforementioned extensions can be captured in \MDHT.
We conclude with a brief summary in Section~\ref{sec:summary}.


%% file: mdht.tex
\section{Metric Dynamic Logic of Here-and-There}\label{sec:mdht}

We start by providing some notation for intervals used in the rest of the paper.
Given $m \in \mathbb{Z} \cup \{-\omega\}$ and $n \in \mathbb{Z} \cup \{\omega\}$, we let
\intervoo{m}{n} stand for the set $\{i \in \mathbb{Z} \mid m < i < n\}$,
\intervco{m}{n}       for         $\{i \in \mathbb{Z} \mid m \leq i < n\}$,
\intervoc{m}{n}       for         $\{i \in \mathbb{Z} \mid m < i \leq n\}$,  and
\intervcc{m}{n}       for         $\{i \in \mathbb{Z} \mid m \leq i \leq n\}$.
For simplicity, in what follows, we refrain from distinguishing intervals and integers from their syntactic representation.
%
\input{./syntax}
\input{./semantics}
\input{./properties}
%

%% file: syntax.tex
\subsection{Syntax}\label{sec:syntax}

Given a set \PV\ of propositional variables (called \emph{alphabet}),
a \emph{metric dynamic formula} $\varphi$ and a \emph{path expression} $\rho$ are mutually defined by the
following pair of grammar rules:

\begin{align*}
  \varphi & ::= \myatom \mid \bot \mid \; \mdeventuallyF{\rho}{I}{\varphi} \; \mid \; \mdalwaysF{\rho}{I}{\varphi} \\
  \rho    & ::= \stp \mid \varphi ? \mid \rho + \rho \mid \rho\mathrel{;}\rho \mid \kstar{\rho} \mid \converse{\rho}\ .
\end{align*}
where $\myatom\in\PV$ is an atom and $I$ is an interval of the form
\intervoo{m}{n} with $m\in\mathbb{Z}\cup\{-\omega\}$ and $n\in\mathbb{Z}\cup\{\omega\}$.

This syntax combines aspects from metric~\citep{koymans90a} and dynamic logic~\citep{pratt76a}.
The metric component manifests itself in the interval attached to the two dynamic modal operators.
Apart from this,
the syntax is similar to the one of Dynamic Logic (\DL;~\citealt{hatiko00a})
but differs in the construction of (atomic) path expressions:
While \DL\ uses a separate alphabet for ``{atomic actions}'',
we follow the approach of~\citeauthor{giavar13a} in \LDL~\citeyearpar{giavar13a} and use
a single alphabet \PV\ along with a single atomic path expression given by
the (transition) constant $\stp \not\in \PV$ (read as ``step'').
Thus, each path expression $\rho$ is a regular expression formed with the constant $\stp$ plus the test construct $\varphi?$
that may refer to propositional atoms in alphabet \PV.

As regards syntactic metric aspects, we allow subindex intervals $I$ to use some forms of closed ends so that we let
\intervcc{m}{n} stand for \intervoo{m{-}1}{n{+}1} provided $m\neq-\omega$ and $n\neq\omega$,
\intervco{m}{n}       for \intervoo{m{-}1}{n    } if       $m\neq-\omega$,                    and
\intervoc{m}{n}       for \intervoo{m    }{n{+}1} if                          $n\neq \omega$.
Also,
we sometimes use the subindices `${}\leq n$' and `${}\geq m$'
as abbreviations of intervals
$\intervoc{-\omega}{n}$ and $\intervco{m}{\omega}$, respectively,
whenever $m\neq-\omega$ and $n\neq\omega$.
On the other hand, when $I=\intervoo{-\omega}{\omega}$, we simply omit the subindex $I$ so that:
\begin{align}\label{def:dht:mdht}
  \deventually{\rho}{\varphi} & \eqdef \mdeventually{\rho}{\intervoo{-\omega}{\omega}}{\varphi} &
      \dalways{\rho}{\varphi} & \eqdef     \mdalways{\rho}{\intervoo{-\omega}{\omega}}{\varphi}.
\end{align}
More generally, we refer to interval-free formulas as \emph{dynamic formulas}.
Finally, we define the inversion of an interval as
\(
\invert{(m..n)} \eqdef ({-n}..{-m})
\).
Note that $I=\invert{(\invert{I})}$ and $i\in I$ iff $-i\in\invert{I}$ for any interval $I$.

As usual,
a \emph{metric dynamic theory} is a (possibly infinite) set of metric dynamic formulas.
We sometimes drop the adjectives and simply use the terms formula and theory,
whenever it is clear from context.

As we show below, the above language allows us to capture several derived operators, like the Boolean ones:
\begin{align}
  \varphi \to    \psi & \eqdef \dalways{\varphi?} \psi &
  \varphi \wedge \psi & \eqdef \deventually{\varphi?} \psi
    \\
  \top & \eqdef \bot \to \bot & \varphi \vee \psi & \eqdef \deventually{\varphi?+\psi?} \top
  \\
  \neg \varphi      & \eqdef \varphi \to \bot
\end{align}
All Boolean connectives $\wedge$, $\vee$, and $\to$ are defined in terms of the essentially dynamic operators
\dalways{\cdot}{} and \deventually{\cdot} without any reference to a restricted underlying interval%
\footnote{%
  No additional expressive power is gained by adding intervals to Boolean operators,
  since they all depend on modalities for test expressions, checked in the current state.
  For instance, $p \to_{\leq 20} q$ would amount to $\dalways{p?}_{\leq 20}\, q$ and
  we see below that this is equivalent to $\dalways{p?} q \equiv p \to q$. Similarly, $p \to_{> 20} q$ would amount to $\top$.}.
Among them, the definition of $\to$ is most noteworthy since it hints at the implicative nature of \dalways{\cdot}.
Negation $\neg$ is then expressed via implication, as usual in \HT.

A formula is \emph{propositional}, if all its connectives are Boolean.
A formula is said to be \emph{conditional} if it contains some occurrence of an atom $\myatom\in\PV$ inside an
operator of form $\mdalways{\cdot}{\cdot}{}$;
it is called \emph{unconditional} otherwise.
Note that formulas with atoms in implication antecedents or negated formulas are also conditional, since they are derived from $\mdalways{\cdot}{\cdot}{}$.
For instance, $\dalways{p?} \bot$ is conditional, and is actually the same as $p \to \bot$ and $\neg p$.

As with \LDL,
we sometimes use a propositional formula $\phi$ as a path expression and let it stand for $(\phi?;\stp)$.
Accordingly, we use the following abbreviations:
\begin{align}\label{def:mdht:formula:path:expression}
  \mdeventually{\phi}{I}{\varphi} & \eqdef  \mdeventually{\phi?;\stp}{I}{\varphi} &
      \mdalways{\phi}{I}{\varphi} & \eqdef      \mdalways{\phi?;\stp}{I}{\varphi}.
\end{align}
This also means that the reading of $\top$ as a path expression amounts to $(\top?;\stp)$
which is just equivalent to $\stp$.
Another abbreviation is the sequence of $n$ repetitions of some expression $\rho$ defined as $\rho^0 \eqdef \top?$ and $\rho^{n+1} \eqdef \rho; \rho^n$.
For instance, $\rho^3$ unfolds to $\rho;\rho;\rho;\top?$ and amounts to $\rho;\rho;\rho$.

\begin{example}\label{ex:sos}
As soon as an accident ($a$) happens onboard, a boat starts sending an SOS ($s$) during the next 10 minutes, while no help ($h$) reply is received.
When the rescue station receives an SOS, it sends back a help reply no later than 2 minutes.
However, the rescue station has only begun working at minute $t=50$.
If an accident happened at time instant $t=40$, would the boat receive a $\mathit{help}$ reply and, if so, when?
\end{example}

A possible formalization of the example could be:
\begin{align}
\alwaysF ( \ a \to \dalways{(\neg h)^*; \neg h}_{\leq 10}\, s \ ) \label{f:sos.1}\\
\alwaysF_{\geq 50} ( s \to \eventuallyF_{\leq 2}\, h ) \label{f:sos.2}\\
\eventuallyF_{40}\, a \label{f:sos.3}
\end{align}

Negation in  \eqref{f:sos.1} acts as a default: we send the SOS while \emph{no evidence about help} is derived.
That formula also combines path expressions from \LDL\ with metric information.


%% file: semantics.tex
\subsection{Semantics}
\label{sec:semantics}

The traditional semantics of linear-time logics, like \LTL~\citep{pnueli77a}, \LDL~\citep{giavar13a}, and \MTL~\citep{koymans90a},
rely on the concept of a \emph{trace},
a (possibly infinite) sequence of \emph{states}, each of which is represented by a set of atoms.
More precisely, a \emph{trace} \Ttrace\ of length $\lambda$ over alphabet \PV\ is
a sequence $\Ttrace=(T_i)_{\rangeco{i}{0}{\lambda}}$ of sets $T_i\subseteq\PV$.
We sometimes use the notation $|\Ttrace| \eqdef \lambda$ to stand for the length of the trace.
We say that \Ttrace\ is \emph{infinite} if $|\Ttrace|=\omega$ and \emph{finite} if $|\Ttrace| \in \mathbb{N}$.
To represent traces, we write a sequence of sets of atoms concatenated with `$\cdot$'.
For instance, the finite trace $\{a\} \cdot \emptyset \cdot \{a\} \cdot \emptyset$ has length 4 and makes $a$ true at even time steps and false at odd ones.
For infinite traces, we sometimes use $\omega$-regular expressions like, for instance,
in the infinite trace $(\{a\} \cdot \emptyset)^\omega$ where all even positions make $a$ true and all odd positions make it false.

At each state $T_i$ in a trace, an atom $a$  can only be true, viz.\ $a \in T_i$, or false, $a \not\in T_i$.
Temporal logics based on the logic of Here-and-There,
like \THT~\citep{agcadipevi13a,agcadipescscvi20a}, \DHT~\citep{bocadisc18a,cadisc19a}, and \MHT~\citep{cadiscsc20a,cadiscsc21a,cadiscsc22a},
weaken such truth assignments,
following the same intuitions as the (non-temporal) logic of \HT:
An atom can have one of three truth-values in each state, namely,
\emph{false}, \emph{assumed} (or true by default) or \emph{proven} (or certainly true).
Anything proved has to be assumed, but the opposite does not necessarily hold.
Following this idea,
a state at time step $i$ is represented as a pair of sets of atoms $\tuple{H_i,T_i}$ with $H_i \subseteq T_i \subseteq \PV$ where
$H_i$ (standing for ``here'') contains the proven atoms, whereas $T_i$ (standing for ``there'') contains the assumed atoms.
On the other hand, false atoms are just the ones not assumed, captured by $\PV \setminus T_i$.
Accordingly,
a \emph{Here-and-There trace} (for short \emph{\HT-trace}) of length $\lambda$ over alphabet \PV\ is
a sequence of pairs
\(
(\tuple{H_i,T_i})_{\rangeco{i}{0}{\lambda}}
\)
with $H_i\subseteq T_i$ for any $\rangeco{i}{0}{\lambda}$.
For convenience, we usually represent the \HT-trace as the pair $\tuple{\Htrace,\Ttrace}$ of traces $\Htrace = (H_i)_{\rangeco{i}{0}{\lambda}}$ and $\Ttrace = (T_i)_{\rangeco{i}{0}{\lambda}}$.
Given $\M=\tuple{\Htrace,\Ttrace}$, we also denote its length as $|\M| \eqdef |\Htrace|=|\Ttrace|=\lambda$.
Note that the two traces \Htrace, \Ttrace\ must satisfy a kind of order relation, since $H_i \subseteq T_i$ for each time step $i$.
Formally, we define the ordering $\Htrace \leq \Ttrace$ between two traces of the same length $\lambda$ as $H_i\subseteq T_i$ for each $\rangeco{i}{0}{\lambda}$.
Furthermore, we define $\Htrace<\Ttrace$ whenever both $\Htrace\leq\Ttrace$ and $\Htrace\neq\Ttrace$.
Thus, an \HT-trace can also be defined as any pair $\tuple{\Htrace,\Ttrace}$ of traces such that $\Htrace \leq \Ttrace$.
The particular type of \HT-traces satisfying $\Htrace=\Ttrace$ are called \emph{total}.

In metric logics, each state is further associated with a \emph{time point}.
For this, we use a function $\tmf$ that assigns to each state index $i \in \intervco{0}{\lambda}$ a time point $\tmf(i)\in \mathbb{N}$
representing the number of time units (seconds, milliseconds, etc, depending on the chosen granularity)
elapsed since time point $\tmf(0)=0$ chosen as the beginning of the trace.
All this leads to the concept of a timed \HT-trace.
%
\begin{definition}\label{def:timed:trace}
  A \emph{timed \HT-trace} over $(\mathbb{N},<)$ is is a triple $\tuple{\Htrace,\Ttrace,\tmf}$ where
  \begin{enumerate}
  \item $\tuple{\Htrace,\Ttrace} = (\tuple{H_i,T_i})_{\rangeco{i}{0}{\lambda}}$ is an \HT-trace and
  \item $\tmf: \intervco{0}{\lambda} \to \mathbb{N}$ is a function
    such that%
    \footnote{This condition correspond to \emph{strict} time traces, which are the ones we use in the current paper.
      Non-strict traces would allow $\tau(i)\leq \tau(i{+}1)$, meaning that, at the same time point, more than one transition could occur,
      even an infinite number of them.
      This seems a less practical possibility, and is left for future study.}
    $\tau(i)<\tau(i{+}1)$,
    \\
    for all $i \in \intervco{0}{\lambda}$ with $i+1 < \lambda$.
  \end{enumerate}
  We assume w.l.o.g.\ that $\tmf(0)=0$.
\end{definition}
%
Given any timed \HT-trace,
satisfaction of formulas is defined as follows.
%
\begin{definition}[\MDHT\ satisfaction]\label{def:mdht:satisfaction}
  A timed \HT-trace $\M=\tuple{\Htrace,\Ttrace,\tmf}$ of length $\lambda$ over alphabet \PV\
  \emph{satisfies} a metric dynamic formula $\varphi$ at time step $\rangeco{k}{0}{\lambda}$,
  written $\M,k \models \varphi$, if the following conditions hold:
  \begin{enumerate}
  \item \label{def:mdhtsat:top:bot}
    $\M,k \not\models \bot$
  \item \label{def:mdhtsat:atom}
    $\M,k \models \myatom$ if $\myatom \in H_k$ for any atom $\myatom \in \PV$
  \item \label{def:mdhtsat:eventually:f}
    $\M, k \models \mdeventuallyF{\rho}{I}{\varphi}$
    if $\M,i \models \varphi$
    for some $i$ with
    $(k,i) \in \mdrel{\rho}{\M}$
    and
    $\tmf(i)-\tmf(k) \in I$
  \item \label{def:mdhtsat:always:f} $\M, k \models \mdalwaysF{\rho}{I}{\varphi}$
    if $\M',i \models \varphi$
    for all $i$ with
    $(k,i) \in \mdrel{\rho}{\M'}$
    and
    $\tmf(i)-\tmf(k) \in I$ for both $\M'=\M$ and $\M'=\tuple{\Ttrace,\Ttrace,\tmf}$
  \end{enumerate}
  where, for any timed \HT-trace $\M$, $\mdrel{\rho}{\M} \subseteq \mathbb{N}^2$ is a relation on pairs of time steps inductively defined as follows.
  \begin{enumerate}
  \setcounter{enumi}{4}
  \item \label{def:mdhtsat:step}
    $\mdrel{\stp}{\M}                    \ \eqdef\ \{ (k,k+1) \mid \rangeco{k,k+1}{0}{\lambda} \}$
  \item \label{def:mdhtsat:test}
    $\mdrel{\varphi?}{\M}                \ \eqdef\ \{ (k,k) \mid  \M,k \models \varphi \}$
  \item \label{def:mdhtsat:ndet}
    $\mdrel{\rho_1\mathrel{+}\rho_2}{\M} \ \eqdef\ \mdrel{\rho_1}{\M} \cup \mdrel{\rho_2}{\M}$
  \item \label{def:mdhtsat:seq}
    \(
    \mdrel{\rho_1\mathrel{;}\rho_2}{\M}  \ \eqdef\ \{ (k,i)  \mid {(k,j) \in \mdrel{\rho_1}{\M}} \text{and }
    {(j,i) \in \mdrel{\rho_2}{\M}} \text{for some } j \}
    \)
  \item \label{def:mdhtsat:star}
    $\mdrel{\kstar{\rho}}{\M}            \ \eqdef\  \bigcup_{n\geq 0} \mdrel{\rho^n}{\M}$
  \item  \label{def:mdhtsat:converse}
    $\mdrel{\converse{\rho}}{\M}         \ \eqdef\ \{ (k,i)\mid (i,k)\in\mdrel{\rho}{\M} \}$
  \end{enumerate}
\end{definition}
%
This definition accommodates traces of arbitrary length $\lambda$,
including both infinite ($\lambda=\omega$) and finite ($\lambda \in \mathbb{N}$) ones.
The most significant feature introduced by the \HT\ part of this semantics is the treatment of the always operator \mdalwaysF{\rho}{I}{\varphi},
which is analogous to the handling of implication\footnote{In fact, this is further inherited from the behavior of implication in intuitionistic logic, whose satisfaction must be checked in all accessible worlds. In this case, the accessibility relation of traces is reflexive and, additionally, \Ttrace\ is accessible from \Htrace.} in the propositional intermediate logic of \HT.
It requires that $\varphi$ is satisfied in ``both dimensions'' \Htrace\ (here) and \Ttrace\ (there) of the trace,
using $\tuple{\Ttrace,\Ttrace,\tmf}$ in addition to $\tuple{\Htrace,\Ttrace,\tmf}$ as with the other connectives.

A formula $\varphi$ is a \emph{tautology} (or is valid), written $\models \varphi$, iff
$\M,k \models \varphi$ for any timed \HT-trace \M\ and any $k \in \intervco{0}{\lambda}$.
\MDHT\ is the logic induced by the set of all such tautologies.
For two formulas $\varphi, \psi$ we write $\varphi \equiv \psi$, iff
$\models \varphi \leftrightarrow \psi$, that is,
$\M,k \models \varphi \leftrightarrow \psi$ for any
timed \HT-trace \M\ of length $\lambda$ and any $k \in \intervco{0}{\lambda}$.
A timed \HT-trace \M\ is an \MDHT\ \emph{model} of a metric dynamic theory $\Gamma$ if $\M,0 \models \varphi$ for all $\varphi \in \Gamma$.
The set of \MDHT\ models of $\Gamma$ having length $\lambda$ is denoted as $\MDHT(\Gamma,\lambda)$,
whereas $\MDHT(\Gamma) \eqdef \bigcup_{\lambda=0}^\omega \MDHT(\Gamma,\lambda)$ is
the set of all \MDHT\ models of $\Gamma$ of any length.
We may obtain fragments of any metric dynamic logic by imposing restrictions on the timed traces used for defining tautologies and models.
We indicate the restriction to finite (resp. infinite) traces by using subscript $f$ (resp. $\omega$).
That is, \MDHTf\ stands for the restriction of \MDHT\ to traces of any finite length $\lambda \in \mathbb{N}$ and
\MDHTo\ corresponds to the restriction to traces of infinite length $\lambda=\omega$.

As we saw, Boolean operators are actually derived.
This allows us to simplify the number of cases to be checked in proofs by structural induction.
Yet, interpreting derived operators by unfolding their definitions can become, in general, a bit cumbersome.
The following proposition provides a direct semantic characterization for Boolean connectives and it can be easily deduced from their definitions as derived operators:
%
\begin{proposition}[Satisfaction of Boolean formulas]\label{prop:satisfaction:bool}
  Let $\M=\tuple{\Htrace,\Ttrace,\tau}$
  be a timed \HT-trace of length $\lambda$ over \PV.
  Given the respective definitions of the Boolean operators, we get the following satisfaction conditions:
  \begin{enumerate}[label={\rm (\roman*)}]
  \item\label{prop:satisfaction:bool:top}
    $\M, k \models \top$
  \item\label{prop:satisfaction:bool:and}
    $\M, k \models \varphi \wedge \psi$
    iff
    $\M, k \models \varphi$
    and
    $\M, k \models \psi$
  \item\label{prop:satisfaction:bool:or}
    $\M, k \models \varphi \vee \psi$
    iff
    $\M, k \models \varphi$
    or
    $\M, k \models \psi$
  \item\label{prop:satisfaction:bool:to}
    $\M, k \models \varphi \to \psi$
    iff
    $\M', k \not \models \varphi$
    or
    $\M', k \models  \psi$, for both $\M'=\M$ and $\M'=\tuple{\Ttrace,\Ttrace,\tmf}$
  \item\label{prop:satisfaction:bool:neg}
    $\M, k \models \neg \varphi$
    iff
    $\tuple{\Ttrace, \Ttrace,\tau}, k \not \models \varphi$
  \end{enumerate}
\end{proposition}


%% file: properties.tex
\subsection{Properties}
\label{sec:properties}

We prove next that a pair of basic properties from \HT\ are maintained in the current extension of \MDHT.
%
\begin{proposition}[Persistence]\label{prop:persistence}
  Let $\tuple{\Htrace,\Ttrace,\tau}$ be a timed \HT-trace of length $\lambda$ over \PV\ and let $\varphi$ be a formula over \PV.

  Then, for any $\rangeco{k}{0}{\lambda}$,
  \begin{enumerate}
  \item\label{prop:persistence:ht} if $\tuple{\Htrace,\Ttrace,\tau}, k \models \varphi$ then $\tuple{\Ttrace,\Ttrace, \tau}, k \models \varphi$ and
  \item\label{prop:persistence:p}  $\mdrel{\rho}{\tuple{\Htrace,\Ttrace,\tau}} \subseteq \mdrel{\rho}{\tuple{\Ttrace,\Ttrace,\tau}}$.

  \end{enumerate}
\end{proposition}
%
Persistence is a property known from intuitionistic logic;
it expresses that accessible worlds satisfy the same or more formulas than the current world,
where \Ttrace\ is ``accessible'' from \Htrace\ in \HT.
This also explains the semantics of $\mdalways{\cdot}{\cdot}{}$,
which behaves as a kind of intuitionistic implication (used to define `$\to$' as a derived operator)
and so, it must hold for all accessible worlds, viz.\ $\tuple{\Htrace,\Ttrace,\tau}$ and $\tuple{\Ttrace,\Ttrace,\tau}$.
%
%
%

%% file: mdel.tex
\section{Metric Dynamic Equilibrium Logic}
\label{sec:mdel}

Given a set of models in \MDHT, we define the ones in equilibrium as follows.
%
\begin{definition}[$\mathfrak{S}$-Equilibrium/$\mathfrak{S}$-Stability]\label{def:tem}
  Let $\mathfrak{S}$ be some set of timed \HT-traces.
  A total timed \HT-trace of the form $\tuple{\Ttrace, \Ttrace,\tau}$ is said to be in \emph{equilibrium} in $\mathfrak{S}$ iff $\tuple{\Ttrace, \Ttrace,\tau} \in\mathfrak{S}$ and
  there is no other $\tuple{\Htrace, \Ttrace,\tau} \in \mathfrak{S}$ such that $\Htrace < \Ttrace$.
When this happens, the timed trace $\tuple{\Ttrace, \tau}$ is said to be  \emph{stable} in $\mathfrak{S}$.
  \qed
\end{definition}
%
When $\mathfrak{S}=\MDHT(\Gamma)$, the timed \HT-traces in equilibrium are called the \emph{metric dynamic equilibrium models} of $\Gamma$, and denoted as $\MDEL(\Gamma)$.
Similarly, $\MDEL(\Gamma,\lambda)$ denotes the \HT-traces in equilibrium for $\MDHT(\Gamma,\lambda)$.
%
%
Since the ordering relation among traces is only defined for a fixed $\lambda$, the following can be easily observed:
%
\begin{proposition}
The set of metric dynamic equilibrium models of a theory $\Gamma$ can be partitioned by the trace length $\lambda$, that is,
$\bigcup_{\lambda=0}^\omega \MDEL(\Gamma,\lambda) = \MDEL(\Gamma)$. \qed
\end{proposition}

\emph{Metric Dynamic Equilibrium Logic} (\MDEL) is the (non-monotonic) logic induced by metric dynamic equilibrium models.
We can also define the variants \MDELo\ and \MDELf\ by applying the corresponding restriction to infinite and finite traces, respectively.

\begin{example}[Example~\ref{ex:sos} continued]
Back to our example, suppose we have the theory $\Gamma$ consisting of the formulas \eqref{f:sos.1}-\eqref{f:sos.3} and let us informally analyze the obtained metric dynamic equilibrium models, or equilibrium models for short.
%
%
%
Let us consider first the possible untimed traces, forgetting the metric subindices by now.
In other words, consider the temporal equilibrium models of:
\begin{align}
\alwaysF ( \ a \to \dalways{(\neg h)^*; \neg h}_{}\, s \ ) \label{f:sos.b.1}\\
\alwaysF_{} ( s \to \eventuallyF_{}\, h ) \label{f:sos.b.2}\\
\eventuallyF_{}\, a \label{f:sos.b.3}
\end{align}
Formula \eqref{f:sos.b.3} forces us to include at least an atom $a$ in some state at time step $i$.
Other total models $\tuple{\Ttrace,\Ttrace,\tau}$ may include more occurrences of $a$ in other states,
but they are not in equilibrium,
since we can always form a smaller model $\tuple{\Htrace,\Ttrace,\tau}$ where those additional $a$'s are removed in $\Htrace$.
This is because, apart from \eqref{f:sos.b.3}, atom $a$ only occurs in the antecedent of the implication in \eqref{f:sos.b.1}.
Now, after some analysis (see~\citep{cadisc19a}), it can be seen that \eqref{f:sos.b.1} is equivalent to the conjunction of temporal rules of the form:
\begin{eqnarray}
\alwaysF ( \ a \wedge \neg h & \to & \wnext s \ ) \label{f:sos.b.1.1}\\
\alwaysF ( \ a \wedge \neg h \wedge \wnext \neg h & \to  & \wnext^2 s \ ) \\
\alwaysF ( \ a \wedge \neg h \wedge \wnext \neg h \wedge \wnext^2 \neg h & \to  & \wnext^3 s \ ) \\
\alwaysF ( \ a \wedge \neg h \wedge \wnext \neg h \wedge \wnext^2 \neg h \wedge \wnext^3 \neg h  & \to  & \wnext^4 s \ ) \\
& \vdots & \nonumber
\end{eqnarray}
As an illustration of their behavior, these rules have a direct correspondence to the ASP program:
\begin{verbatim}
s(T+1) :- a(T), not h(T), step(T+1).
s(T+2) :- a(T), not h(T), not h(T+1), step(T+2).
s(T+3) :- a(T), not h(T), not h(T+1), not h(T+2), step(T+3).
s(T+4) :- a(T), not h(T), not h(T+1), not h(T+2), not h(T+3), step(T+4).
...
\end{verbatim}
where we use a time argument for all atoms and a predicate {\tt step} to describe the existing time points from $0$ to $|\Ttrace|-1$.
Since $a$ only holds at state $i$, the first possible justified occurrence of $s$ could only appear from $i+1$ on.
This means that any occurrence of $s$ up to $i$ can be removed in a smaller $\Htrace$ because, apart from \eqref{f:sos.b.1}, $s$ only occurs in the antecedent of \eqref{f:sos.b.2}.
But this allows us to conclude that there is no evidence for $h$ at $i$, since \eqref{f:sos.b.2} has a false antecedent at that state.
In other words, any equilibrium model begins with a prefix $\emptyset^* \cdot \{a\}$.
Implication \eqref{f:sos.b.1.1} at $i$ forces the existence of a state $i+1$ where $s$ must hold, but then, \eqref{f:sos.b.2} allows us to conclude $\eventuallyF h$ at $i+1$.
By a similar reason as we followed for \eqref{f:sos.b.3}, we conclude that $h$ is true at exactly one state $j \geq i+1$.
Since all the states between $i+1$ and $j-1$ satisfy $\neg h$, the boat keeps sending an SOS due to \eqref{f:sos.b.1}.
To sum up, the possible temporal equilibrium models have one of the forms:
\begin{eqnarray*}
&&\emptyset^* \cdot \{a\}  \\
&&\emptyset^* \cdot \{a\} \cdot \{s\}^* \cdot \{s,h\} \cdot \emptyset^*
\end{eqnarray*}
where, we use $T^*$ to mean any (possibly empty) sequence of repetitions of state $T$.
The first case corresponds to finite-length stable models where the accident just happens at the end of the trace.
In the second case, when the boat receives the help reply, it still sends a last SOS since in the previous state, $h$ was false yet.

Now, let us see what happens when we consider the metric information.
Formula \eqref{f:sos.3} fixes the time point $\tau(i)=40$ for the state $i$ at which $a$ is made true.
Implication \eqref{f:sos.1} is weaker now, since it is only applicable if the states in which we send the SOS are 10 minutes ago from the accident at $i$.
Further than that, the boat stops sending messages.
One new possible outcome, for instance, is that there is no transition at all within the next 10 minutes, and so, no SOS is ever emitted.
This leads to equilibrium models of the form $\emptyset^* \cdot \{a\}$ (when the accident is at the last step) or the form
\[
\xymatrix @-5mm {
\emptyset^* & \cdot & \{a\} & \cdot & \emptyset & \cdot & \emptyset^*\\
& & i \ar[u] & & i+1 \ar[u] & 
}
\]
for the state $i$ where $a$ occurs, with $\tau(i)=40$ and $\tau(i+1)>50$.
If, on the contrary, there is at least one transition in the next 10 minutes, we must note now that \eqref{f:sos.2} is only effective now at points $j$ where $\tau(j)\geq 50$.
So, the rescue station does not attend to any signal before $t=50$.
A second possibility is, therefore, that the boat sends one or more SOS calls, but at transitions that do not reach moment $t=50$.
This would correspond to equilibrium models of the form
\[
\xymatrix @-5mm {
\emptyset^* & \cdot & \{a\} & \cdot & \{s\}^* & \cdot & \{s\} & \cdot & \emptyset^*\\
& & i \ar[u] & & & & j \ar[u] &
}
\]
where, $j>i$, $\tau(i)=40$ and $\tau(j)<50$ and the SOS calls are never attended.
Finally, it may be the case that the last SOS call within the 10 minutes interval happens exactly at $t=50$, when the rescue station begins operating.
If so, a help reply must be sent within the next 2 minutes.
This final group of equilibrium models would have either the form:
\[
\xymatrix @-5mm {
\emptyset^* & \cdot & \{a\} & \cdot & \{s\}^* & \cdot & \{s,h\} & \cdot & \emptyset^*\\
& & i \ar[u] & & & & j \ar[u] &
}
\]
with $\tau(i)=40$ and $\tau(j)=50$, meaning that the help is immediate, or perhaps:
\[
\xymatrix @-5mm {
\emptyset^* & \cdot & \{a\} & \cdot & \{s\}^* & \cdot & \{s\} & \cdot & \emptyset^* & \cdot & \{h\} & \cdot & \emptyset^*\\
& & i \ar[u] & & & & j \ar[u] & & & & k \ar[u]
}
\]
\end{example}
where $\tau(i)=40$, $\tau(j)=50$ and $\tau(k)\leq 52$.
%

%% file: relation.tex
\section{Relationships}
\label{sec:relation}

In this section we show that \MDHT\ constitutes an umbrella where different logics for temporal ASP,
most of them previously introduced in the literature, can be simply identified as fragments.

\subsection{Metric Dynamic Logic}
\label{sec:mdl}

An interesting subset of \MDHT\ is the one formed by total timed traces $\tuple{\Ttrace, \Ttrace,\tau}$.
In the non-metric version of temporal and dynamic \HT, viz.\ \THT\ and \DHT, the restriction to
total models corresponds to Linear Temporal and Dynamic Logic (\LTL\ and \LDL~\citep{pnueli77a,giavar13a}).
In our case, the restriction to total traces defines a metric version of \LDL,
that we simply call \emph{Metric Dynamic Logic} (\MDL\ for short).
\footnote{This designation is somewhat justified in view of Corollary~\ref{cor:mdl:mdht}.}

For simplicity, we refrain from an explicit introduction of the semantics of \MDL,
since it just corresponds to \MDHT\ on total traces $\tuple{\Ttrace,\Ttrace,\tmf}$, as stated below.
Let us simply use $\tuple{\Ttrace,\tmf},k \models \varphi$ to denote the satisfaction of $\varphi$ by a timed trace $\tuple{\Ttrace,\tmf}$ at point $k$ in \MDL\
and \mdrel{\rho}{\tuple{\Ttrace,\tmf}} the \MDL\ accessibility relation for $\rho$ and $\tuple{\Ttrace,\tmf}$.
In fact, $\tuple{\Ttrace,\tmf},k \models \varphi$ is obtained from $\M,k \models \varphi$ in
Definition~\ref{def:mdht:satisfaction} by replacing \M\ by \tuple{\Ttrace,\tmf} and by dropping
Condition~\ref{def:mdhtsat:always:f} and rather defining $\M, k \models \mdalwaysF{\rho}{I}{\varphi}$ as
 $\M, k \models \neg\mdeventuallyF{\rho}{I}{\neg\varphi}$.
%
\begin{proposition}\label{prop:totality}
  For any total timed \HT-trace $\tuple{\Ttrace,\Ttrace,\tau}$ of length $\lambda$,
  any formula $\varphi$ and any path expression $\rho$,
  we have that
  \begin{enumerate}
  \item $\tuple{\Ttrace,\Ttrace,\tau},k \models \varphi$ iff $\tuple{\Ttrace,\tau},k \models \varphi$, for all $\rangeco{k}{0}{\lambda}$; and
  \item $\mdrel{\rho}{\tuple{\Ttrace,\Ttrace,\tau}} = \mdrel{\rho}{\tuple{\Ttrace,\tau}}$.
  \end{enumerate}
\end{proposition}
%
Accordingly, any total timed \HT-trace $\tuple{\Ttrace,\Ttrace,\tau}$ can be seen as the timed trace $\tuple{\Ttrace,\tau}$.
In fact, under total models, the satisfaction of dynamic operators \mdeventually{\cdot}{\cdot} and \mdalways{\cdot}{\cdot} in \MDHT\ collapses to that in \MDL.

This gives rise to the following result.
%
\begin{proposition}\label{prop:no:implication}
  Let $\varphi$ and $\psi$ be unconditional,\footnote{That is, formulas without implication, and so, without negation either.} metric dynamic formulas.
  Then, $\varphi \equiv \psi$ in \MDL\ iff $\varphi \equiv \psi$ in \MDHT.
\end{proposition}

Moreover, the first item (along with Proposition~\ref{prop:persistence}) implies that any \MDHT\ tautology is also an \MDL\ tautology, so the former constitutes a weaker logic.
To show that, in fact, \MDHT\ is \emph{strictly} weaker,
note that it does not satisfy some classical tautologies like the \emph{excluded middle} $\varphi\vee\neg\varphi$,
while \MDL\ is a proper extension of classical logic.
In fact, the addition of the axiom schema
\begin{align}
\alwaysF (a \vee \neg a) \tag{\EM}
\quad
\text{ for each atom }
a \in \mathcal{A}
\text{ in the alphabet}
\end{align}
forces total models and so, makes \MDHT\ collapse to \MDL.
Propositions~\ref{prop:persistence} and~\ref{prop:totality} imply that $\varphi$ is \MDHT\ satisfiable iff it is \MDL\ satisfiable.

In this context, it may also be interesting to recall item \ref{prop:satisfaction:bool:neg} in Proposition~\ref{prop:satisfaction:bool},
showing that the satisfaction of negated formulas amounts to checking the formulas' dissatisfaction in the total trace obtained from the `there' world,
namely,
$\tuple{\Htrace,\Ttrace,\tau}, k \models \neg \varphi$
iff
$\tuple{\Ttrace, \Ttrace,\tau}, k \not \models \varphi$ iff $\tuple{\Ttrace,\tau}, k \not \models \varphi$.
In other words,
the satisfaction of negated formulas amounts to that in \MDL.

In fact,
the metric dynamic logic defined in~\citep{bakrtr17a} corresponds to a fragment of \MDHT\ when restricting ourselves to total traces.
Metric dynamic formulas are defined in~\citep{bakrtr17a} over intervals of natural numbers using negation, disjunction as well as a future- and
past-oriented eventually operator.
The latter can be expressed in \MDHT\ by applying the converse operator to the path expression and by inverting the interval,
viz.\ \mdeventuallyF{\converse{\rho}}{\invert{I}}{\varphi}.

In the next result, we use $\varphi^-$ to denote the result of replacing accordingly in formula $\varphi$
each subformula with such a past-oriented eventually operator.
%
\begin{corollary}\label{cor:mdl:mdht}
  Let $\varphi$ be a metric dynamic formula restricted to the fragment used in~\citep{bakrtr17a}, $\tuple{\Ttrace,\tmf}$ a timed trace and $k \geq 0$.

  Then, $\tuple{\Ttrace,\tmf},k \models \varphi$ under \MDL\ satisfaction as in~\citep{bakrtr17a} iff $\tuple{\Ttrace,\Ttrace},k \models \varphi^-$ under \MDHT\ satisfaction.
\end{corollary}

\subsection{The Dynamic Logic of Here-and-There and Dynamic Equilibrium Logic}
\label{sec:dht}

We consider now the (non-metric) linear dynamic extension of \HT\ introduced in~\citep{cadisc19a} known as \DHT.
A formula is called \emph{dynamic},
if all its intervals equal \intervoo{-\omega}{\omega},
in which case we also drop all intervals.
%
\begin{proposition}[\DHT-satisfaction of dynamic formulas]\label{prop:satisfaction:dht}
  Let $\M=\tuple{\Htrace,\Ttrace,\tau}$
  be a timed \HT-trace of length $\lambda$ over \PV.
  Given the respective definitions of derived operators, we get the following satisfaction conditions:
  \begin{enumerate}
  \item \label{def:dhtsat:eventually}
    $\M, k \models \deventually{\rho}{\varphi}$
    iff $\M,i \models \varphi$
    for some $i$ with $(k,i) \in \mdrel{\rho}{\M}$
  \item \label{def:dhtsat:always}
    $\M, k \models \dalways{\rho}{\varphi}$
    iff $\M',i \models \varphi$
    for all $i$ with $(k,i) \in \mdrel{\rho}{\M'}$ \\ for both $\M'=\M$ and $\M'=\tuple{\Ttrace,\Ttrace,\tau}$
  \end{enumerate}
  where $\mdrel{\rho}{\M}$ is defined as in Definition~\ref{def:mdht:satisfaction}.
\end{proposition}

Note that, for the syntactic fragment dynamic formulas, the time function $\tmf$ is never used and can be dropped.
As a consequence:
\begin{corollary}\label{cor:dht:mdht}
  Let $\varphi$ be a dynamic formula and \M\ a timed \HT-trace and $k \geq 0$.

  Then, $\M,k \models \varphi$ under \DHT\ satisfaction iff $\M,k \models \varphi$ under \MDHT\ satisfaction.
\end{corollary}
and, since $\THT$ can be defined as a fragment of $\DHT$ we also obtain:
\begin{corollary}\label{cor:dht:tht}
	Let $\varphi$ be a temporal formula and \M\ a timed \HT-trace and $k \geq 0$.

	Then, $\M,k \models \varphi$ under \THT\ satisfaction iff $\M,k \models \varphi$ under \MDHT\ satisfaction.
\end{corollary}

Since the models selection of \MDEL\ for dynamic and temporal formulas respectively collapse to the model selection of
\DEL\ and \TEL, we can directly conclude:
%
\begin{corollary}
Let $\varphi$ be a dynamic (resp.\ temporal) formula.
A total timed trace $\tuple{\Ttrace, \Ttrace,\tau}$ is a metric dynamic equilibrium model of $\varphi$ under $\MDEL$ definition iff it is a dynamic (resp.\ temporal) equilibrium logic of $\varphi$ under $\DEL$ (resp.\ \TEL) definition.
\end{corollary}

\subsection{The Metric Logic of Here-and-There and Metric Equilibrium Logic}
\label{sec:mht}

Similar to the expression of Boolean operators in terms of (metric) dynamic formulas,
we can also express metric and temporal formulas.

Given an interval $I$,
operators \mdalways{\cdot}{I} and \mdeventually{\cdot}{I} allow us also to define future metric operators
\finally,
\metricI{\next},
\metricI{\wnext},
\metricI{\eventuallyF},
\metricI{\alwaysF},
\metricI{\until},
standing for
\emph{final,
next,
weak next,
eventually,
always,
until,}
and their past-oriented counterparts:
\initially,
\metricI{\previous},
\metricI{\wprevious},
\metricI{\eventuallyP},
\metricI{\alwaysP},
\metricI{\since}.
%

\begin{align}
  \label{def:mht:final:initial}
  \finally                        & \eqdef \dalways{\stp}{\bot}                            & \initially                     & \eqdef \dalways{\converse{\stp}}{\bot}
  \\\label{def:mht:next:previous}
  \metricI{\next} \varphi         & \eqdef \mdeventually{\stp}{I}{\varphi}                 & \metricI{\previous} \varphi    & \eqdef \mdeventually{\converse{\stp}}{\invert{I}}{\varphi}
  \\\label{def:mht:wnext:wprevious}
  \metricI{\wnext} \varphi        & \eqdef \mdalways{\stp}{I}{\varphi}                     & \metricI{\wprevious} \varphi   & \eqdef \mdalways{\converse{\stp}}{\invert{I}}{\varphi}
  \\\label{def:mht:eventually}
  \metricI{\eventuallyF} \varphi  & \eqdef \mdeventually{\kstar{\stp}}{I}{\varphi}         & \metricI{\eventuallyP} \varphi & \eqdef \mdeventually{\converse{\kstar{\stp}}}{\invert{I}}{\varphi}
  \\\label{def:mht:always}
  \metricI{\alwaysF} \varphi      & \eqdef \mdalways{\kstar{\stp}}{I}{\varphi}             & \metricI{\alwaysP} \varphi     & \eqdef \mdalways{\converse{\kstar{\stp}}}{\invert{I}}{\varphi}
  \\\label{def:mht:until:since}
  \varphi \metricI{\until} \psi   & \eqdef \mdeventually{\kstar{(\varphi?;\stp)}}{I}{\psi} & \varphi \metricI{\since} \psi  & \eqdef \mdeventually{\converse{\kstar{(\stp;\varphi?)}}}{\invert{I}}{\psi}
\end{align}
This syntax essentially coincides with the metric (timed) extension of \HT\ presented in~\citep{cadiscsc22a} called \MHT.
A formula is called \emph{metric}, if it includes only Boolean and metric operators.
A metric formula is called \emph{temporal},
if furthermore all its intervals equal $\intervoo{-\omega}{\omega}$,
in which case we drop them once more.

Note that \emph{initial} and \emph{final} only depend on the state of the trace and
not on the actual time that this state is mapped to.
Hence, they are not indexed by any interval.

The weak one-step operators, \wnext\ and \wprevious, are of particular interest when dealing with finite traces,
since their behavior differs from their genuine counterparts only at the ends of a trace.
In fact, $\metricI{\wnext} \varphi$ can also be expressed as $\metricI{\next} \varphi \vee \neg \metricI{\next} \top$ (and $\metricI{\wprevious}$ as $\metricI{\previous} \varphi \vee \neg \metricI{\previous} \top$)
However, it can no longer be defined in terms of \emph{final},
as done in~\citep{cakascsc18a} with non-metric $\wnext \varphi \equiv \next \varphi \vee \finally$
(the same applies to weak \emph{previous} and \emph{initial}).

The \MDHT\ definition of $\metricI{\release}$ and $\metricI{\trigger}$ is not as immediate as for the other \MHT\ operators, since they cannot be directly defined in terms of the remaining connectives.
In \LTL, it is well-known that
the formulas $\varphi \release \psi$ (resp.\ $\varphi \trigger \psi$) and $\neg \left( \neg \varphi \until \neg \psi \right)$
(resp.\ $\neg \left( \neg \varphi \since \neg \psi \right)$) are equivalent.
Consequently, $\varphi \release \psi$ corresponds to the \LDL{} expression $\dalways{(\neg \varphi?; \stp)^*}\psi$ while
$\varphi \trigger \psi$ corresponds to $\dalways{(\neg \varphi?;\stp^-)^*}\psi$.
When we move to \DHT, those previous equivalences do not hold any more
because of the constructive behavior of the negation but, yet, release and trigger can still be defined as derived operators thanks to the equivalences~\citep{bocadisc18a}:
\begin{eqnarray*}
\varphi \release \psi \eqdef (\psi \until (\varphi \wedge \psi)) \vee \alwaysF \psi &\quad \quad&
\varphi \trigger \psi  \eqdef (\psi \since (\varphi \wedge \psi)) \vee \alwaysP \psi.
\end{eqnarray*}
Unfortunately, once we add the intervals, it is easy to see that:
\begin{eqnarray*}
  \varphi \metricI{\release} \psi \not \equiv (\psi \metricI{\until} (\varphi \wedge \psi)) \vee \metricI{\alwaysF} \psi &\text{ and }&
                                                                                                                                     \varphi \metricI{\trigger} \psi  \not \equiv (\psi \metricI{\since} (\varphi \wedge \psi)) \vee \metricI{\alwaysP} \psi.
\end{eqnarray*}
To put a counter example for the $\release_I$ equivalence (the one for $\trigger_I$ follows a similar reasoning),  consider the timed \HT-trace $\tuple{\Htrace,\Ttrace,\tmf}$ of length $\lambda=3$,
where $T_0 = \lbrace a \rbrace$, $T_1 = T_2 = \emptyset$, $\tmf(0) = 0$, $\tmf(1) = 1$ and $\tmf(2) = 4$.
It can be easily checked that $\tuple{\Htrace,\Ttrace,\tmf}, 0 \models a \release_{[3,5]} b$ but
$\tuple{\Htrace,\Ttrace,\tmf}, 0 \not \models b \until_{[3,5]} (a \wedge b)$ and $\tuple{\Htrace,\Ttrace,\tmf}, 0 \not \models \alwaysF_{[3,5]} b$.

In \citep{becadiscsc23} it is shown that $\metricI{\release}$ (resp. $\metricI{\trigger}$) can be defined in terms of the unary timed operators and the (untimed) \release\ (resp. \trigger), which are \MDHT-definable.

\begin{align*}
  \varphi \release_{\intervco{m}{n}} \psi &\equiv \alwaysF_{\intervco{m}{n}} \psi \vee \eventuallyF_{\intervco{0}{m}} \left( \varphi \release \left(\varphi \vee \wnext \psi\right)\right) & \varphi \release_{\intervco{0}{n}} \psi &\equiv \alwaysF_{\intervco{0}{n}}\psi \vee \varphi \release \psi \\
  \varphi \release_{\intervcc{m}{n}} \psi &\equiv \alwaysF_{\intervcc{m}{n}} \psi \vee \eventuallyF_{\intervco{0}{m}} \left( \varphi \release \left(\varphi \vee \wnext \psi\right)\right) & \varphi \release_{\intervcc{0}{n}} \psi &\equiv  \alwaysF_{\intervcc{0}{n}}\psi \vee \varphi \release \psi\\
  \varphi \release_{\intervoo{m}{n}} \psi &\equiv \alwaysF_{\intervoo{m}{n}} \psi \vee \eventuallyF_{\intervcc{0}{m}} \left( \varphi \release \left(\varphi \vee \wnext \psi\right)\right) & \varphi \release_{\intervoo{0}{n}} \psi &\equiv  \alwaysF_{\intervoo{0}{n}}\psi \vee \varphi \release \left(\varphi \vee \wnext \psi\right)\\
  \varphi \release_{\intervoc{m}{n}} \psi &\equiv \alwaysF_{\intervoc{m}{n}} \psi \vee \eventuallyF_{\intervcc{0}{m}} \left( \varphi \release \left(\varphi \vee \wnext \psi\right)\right) & \varphi \release_{\intervoc{0}{n}} \psi &\equiv  \alwaysF_{\intervoc{0}{n}}\psi  \vee \varphi \release \left(\varphi \vee \wnext \psi\right)\\
  \varphi \trigger_{\intervco{m}{n}} \psi &\equiv \alwaysP_{\intervco{m}{n}} \psi \vee \eventuallyP_{\intervco{0}{m}} \left( \varphi \trigger \left(\varphi \vee \wprevious \psi\right)\right)&			\varphi \trigger_{\intervco{0}{n}} \psi &\equiv \alwaysP_{\intervco{0}{n}}\psi \vee \varphi \trigger \psi\\
  \varphi \trigger_{\intervcc{m}{n}} \psi &\equiv \alwaysP_{\intervcc{m}{n}} \psi \vee \eventuallyP_{\intervco{0}{m}} \left( \varphi \trigger \left(\varphi \vee \wprevious \psi\right)\right)&			\varphi \trigger_{\intervcc{0}{n}} \psi &\equiv  \alwaysP_{\intervcc{0}{n}}\psi \vee \varphi \trigger \psi\\
  \varphi \trigger_{\intervoo{m}{n}} \psi &\equiv \alwaysP_{\intervoo{m}{n}} \psi \vee \eventuallyP_{\intervcc{0}{m}} \left( \varphi \trigger \left(\varphi \vee \wprevious \psi\right)\right)&			\varphi \trigger_{\intervoo{0}{n}} \psi &\equiv  \alwaysP_{\intervoo{0}{n}}\psi \vee \varphi \trigger \left(\varphi \vee \wprevious \psi\right)\\
  \varphi \trigger_{\intervoc{m}{n}} \psi &\equiv \alwaysP_{\intervoc{m}{n}} \psi \vee \eventuallyP_{\intervcc{0}{m}} \left( \varphi \trigger \left(\varphi \vee \wprevious \psi\right)\right)&			\varphi \trigger_{\intervoc{0}{n}} \psi &\equiv  \alwaysP_{\intervoc{0}{n}}\psi  \vee \varphi \trigger \left(\varphi \vee \wprevious \psi\right)\\
\end{align*}

From the unary normalform it is quite forward to get the respective MDHT translation, since $\release$ and $\trigger$ are definable according to \citep{bocadisc18a}.

Finally, note that the converse operator $\converse{\rho}$ is essential for expressing all metric past operators,
whose addition in temporal logic is exponentially more succinct than using only future operators~\citep{agcadipevi17a}.

%
\begin{proposition}[\MHT-satisfaction of metric temporal logic]\label{prop:satisfaction:mht}
  Let $\M=\tuple{\Htrace,\Ttrace,\tau}$
  be a timed \HT-trace of length $\lambda$ over \PV.
  Given the respective definitions of the metric temporal operators, we get the following satisfaction conditions of \MHT{}:
  \begin{enumerate}
  \item \label{def:mhtsat:initial}
    $\M, k \models \initially$
    iff
    $k =0$
  \item \label{def:mhtsat:previous}
    $\M, k \models \metricI{\previous}\, \varphi$
    iff
    $k>0$ and $\M, k{-}1 \models \varphi$ and $\tmf(k)-\tmf(k{-}1) \in I$
  \item \label{def:mhtsat:wprevious}
    $\M, k \models \metricI{\wprevious}\, \varphi$
    iff
    $k =0$ or
    $\M, k{-}1 \models \varphi$
    or $\tmf(k)-\tmf(k{-}1) \not\in I$
  \item \label{def:mhtsat:eventually:p}
    $\M, k \models \metricI{\eventuallyP}\, \varphi$
    iff
    $\M, i \models \varphi$ for some $\rangecc{i}{0}{k}$
    with
    $\tmf(k)-\tmf(i) \in I$
  \item \label{def:mhtsat:always:p}
    $\M, k \models \metricI{\alwaysP}\, \varphi$
    iff
    $\M, i \models \varphi$ for all $\rangecc{i}{0}{k}$
    with
    $\tmf(k)-\tmf(i) \in I$
  \item \label{def:mhtsat:since}
    $\M, k \models \varphi \metricI{\since} \psi$
    iff
    for some $\rangecc{j}{0}{k}$
    with
    $\tmf(k)-\tmf(j) \in I$,
    we have
    $\M, j \models \psi$
    and
    $\M, i \models \varphi$ for all $\rangeoc{i}{j}{k}$
  \item \label{def:mhtsat:trigger}
    $\M, k \models \varphi \metricI{\trigger} \psi$
    iff
    for all $\rangecc{j}{0}{k}$
    with
    $\tmf(k)-\tmf(j) \in I$,
    we have
    $\M, j \models \psi$
    or
    $\M, i \models \varphi$ for some $\rangeoc{i}{j}{k}$
  \item  \label{def:mhtsat:final}
    $\M, k \models \finally$
    iff
    $k+1 = \lambda$
  \item \label{def:mhtsat:next}
    $\M, k \models \metricI{\next}\, \varphi$
    iff
    $k+1<\lambda$ and $\M, k{+}1 \models \varphi$
    and $\tmf(k{+}1)-\tmf(k) \in I$
  \item \label{def:mhtsat:wnext}
    $\M, k \models \metricI{\wnext}\, \varphi$
    iff
    $k+1=\lambda$
    or $\M, k{+}1 \models \varphi$
    or $\tmf(k{+}1)-\tmf(k) \not\in I$
  \item \label{def:mhtsat:eventually:f}
    $\M, k \models \metricI{\eventuallyF}\, \varphi$
    iff
    $\M, i \models \varphi$ for some $\rangeco{i}{k}{\lambda}$
    with
    $\tmf(i)-\tmf(k) \in I$
  \item \label{def:mhtsat:always:f}
    $\M, k \models \metricI{\alwaysF}\, \varphi$
    iff
    $\M, i \models \varphi$ for all $\rangeco{i}{k}{\lambda}$
    with
    $\tmf(i)-\tmf(k) \in I$
  \item \label{def:mhtsat:until}
    $\M, k \models \varphi \metricI{\until} \psi$
    iff
    for some $\rangeco{j}{k}{\lambda}$
    with
    $\tmf(j)-\tmf(k) \in I$,
    we have
    $\M, j \models \psi$
    and
    $\M, i \models \varphi$ for all $\rangeco{i}{k}{j}$
  \item \label{def:mhtsat:release}
    $\M, k \models \varphi \metricI{\release} \psi$
    iff
    for all $\rangeco{j}{k}{\lambda}$
    with
    $\tmf(j)-\tmf(k) \in I$,
    we have
    $\M, j \models \psi$
    or
    $\M, i \models \varphi$ for some $\rangeco{i}{k}{j}$
    \qed
  \end{enumerate}
\end{proposition}
%
\begin{corollary}
  Let $\varphi$ be a metric formula, \M\ a timed \HT-trace and $k \geq 0$.

  Then, $\M,k \models \varphi$ under \MHT\ satisfaction iff $\M,k \models \varphi$ under \MDHT\ satisfaction.
\end{corollary}

Since the models selection of \MDEL\ for this syntactic fragment collapses to the model selection of \MEL, we can directly conclude:
\begin{corollary}
Let $\varphi$ be a metric formula.
A total timed trace $\tuple{\Ttrace, \Ttrace,\tau}$ is a metric dynamic equilibrium model of $\varphi$ (under $\MDEL$ definition) iff it is a metric equilibrium model of $\varphi$ (under $\MEL$ definition).
\end{corollary}

%% file: discussion.tex
\section{Conclusions}\label{sec:summary}

An important first step towards a logical formalization of temporal ASP systems was initiated with the linear-time temporal extension of Equilibrium Logic and its monotonic basis, the logic of Here-and-There (\HT),  respectively called \emph{Temporal Equilibrium Logic} (\TEL) and \emph{Temporal Here-and-There} (\THT) \citep{agcadipescscvi20a}.
This extension allowed defining temporal stable models for arbitrary theories for the syntax of the well-known Linear-Time Temporal Logic (\LTL)~\citep{pnueli77a} and gave rise to the implementation of the temporal ASP tool \telingo~\citep{cakamosc19a,cadilasc20a}.
In our previous work, we explored extending the \LTL\ syntax in two different directions.
On the one hand, we studied the use of path expressions from Dynamic Logic~\citep{hatiko00a}, adopting the syntax and principles of Linear Dynamic Logic (\LDL)~\citep{giavar13a}.
This led to the extensions called \emph{Dynamic \HT} (\DHT) and \emph{Dynamic Equilibrium Logic} (\DEL) that,
despite of allowing a richer syntax and expressiveness,
used the same semantic structures as \LTL, since a temporal stable model is just a (finite or infinite) trace.
On the other hand, in a somehow orthogonal way, we also studied the incorporation in \THT\ of metric information, by extending modal operators with time intervals as in Metric Temporal Logic (\MTL)~\citep{koymans90a}, leading to \emph{Metric \HT} (\MHT) and \emph{Metric Equilibrium Logic} (\MEL)~\citep{cadiscsc22a}.
This meant not only a syntax extension with respect to \LTL\ operators, but also a generalization of the semantic structure of a temporal stable model, that had in this case the form of a \emph{timed} trace, that is, a trace where each state has an associated time point from an external clock.

In this paper, we have presented a unification of both extensions, namely, the use of dynamic operators and path expressions with the extension for metric intervals and timed traces.
The new formalism, we call \emph{Metric Dynamic Here-and-There} (\MDHT) and its non-monotonic version \emph{Metric Dynamic Equilibrium Logic} (\MDEL), acts as an umbrella for all the mentioned previous logics, that become now fragments of the new approach.
One interesting feature is that, as happens with \DHT, both future and past operators are accommodated into the two standard dynamic modalities, $\dalways{\cdot}_I$ and $\deventually{\cdot}_I$ that have now a metric interval $I$ as a subindex.
This is achieved by a combination of the converse operator $\rho^-$ in path expressions with the use of negative values in the interval $I$.
To put an example, the usual encoding of the (non-metric) past temporal formula $\eventuallyP p$ in \LDL\ is $\deventually{(\stp^*)^-}{} p$, using the converse operator.
If we add a metric interval, as in $\eventuallyP_{\intervcc{2}{9}} p$, then the encoding into \MDHT\ also reverses the interval ends (switching their roles and their signs) leading to $\deventually{(\stp^*)^-}_{\intervcc{-9}{-2}} \, p$.
This extension of intervals with negative numbers has been mostly thought for encoding standard past temporal operators, but also offers new expressive flexibility.
For instance, we can now deal with quite complex and compact formulas like:
\[
\alwaysF \ ( \ \eventuallyP \deventually{(a;b)^*}_{\intervcc{-5}{5}} \, p \to q \ )
\]
meaning that it is always true that if there was some point in the past where $p$ held in the middle of an alternation of $a$ and $b$ that occurred between 5 minutes before and five minutes after $p$, then $q$ must hold now.

For future work, we plan to study different properties of the new formalism, such as strong equivalence, reduction to normal forms, translations of \MDHT\ and \MDEL\ respectively into First and Second Order Logic, or encodings into logic programming.
%
%
Another important topic is exploring the relation with ASP with Constraint Solving~\citep{baboge05a}, and exploiting its use for implementation purposes.


%% file: acknowledgments.tex
\subsection*{Acknowledgments}

This work was supported by
MICINN, Spain, grant PID2020-116201GB-I00, Xunta de Galicia, Spain (GPC ED431B 2019/03), Fundaci{\'o}n BBVA (LIANDA research project), Spain,
R{\'e}gion Pays de la Loire, France (EL4HC and {\'e}toiles montantes CTASP),
DFG grant SCHA 550/15, Germany,
and
European Union COST action CA-17124.


%% file: proofs.tex
\newpage
\section{Auxiliary results}
\label{sec:lemmas}

\input{./proofs/lemmas}
\newpage
\section{Proofs}
\label{sec:proofs}

\input{./proofs/prop-satisfaction-bool}
\input{./proofs/prop-persistence}
\input{./proofs/prop-totality}
\input{./proofs/prop-implications}
\input{./proofs/prop-satisfaction-dht}
\input{./proofs/prop-satisfaction-mht}
\input{./proofs/prop-satisfaction-tht}


%% file: proofs/lemmas.tex
\begin{definition}[Well-founded subexpression~\cite{hatiko00a}]
  In Dynamic Logic, an \emph{expression} can be either a path expression or a formula.
  Either one can be a subexpression of the other because of the mixed operators $\metricI{\dalways{~}}$, $\metricI{\deventually{~}}$ and the test $?$.
\end{definition}

\begin{remark}\label{remark:inecuations} For all $a,b,c \in \mathbb{Z}$, the following relations hold:
  \begin{itemize}
  \item $a-b < c$ iff $b-a > -c$
  \item $a-b \le c$ iff $b-a \ge -c$
  \item $a-b > c$ iff $b-a < -c$
  \item $a-b \le - c$ iff $b-a \ge c$
  \end{itemize}
\end{remark}

\begin{proposition}\label{prop:aux:ref}
  For all $x \in \mathbb{N}$ and for all traces $\M$, $\drel{\stp^x}{\M} = \lbrace  (a,a+x)\rbrace$
\end{proposition}

\begin{proposition}\label{prop:aux:until}
  For all formulas $\varphi$ and for all traces $\M$,
  \[
  \drel{(\varphi?;\stp)^*}{\M} = \lbrace (a,b) \mid a \le b \text{ and } \M, i \models \varphi \text{ for all } a \le i<b \rbrace.
  \]
\end{proposition}

\begin{proposition}\label{prop:aux:since}
  For all formulas $\varphi$ and for all traces $\M$,
  \[
    \drel{(\varphi?;\stp^-)^*}{\M} = \lbrace (a,b) \mid a \ge b \text{ and } \M, i \models \varphi \text{ for all } b<i \le a \rbrace.
    \]
\end{proposition}

%% file: proofs/prop-satisfaction-bool.tex
\begin{proof}\Proposition{prop:satisfaction:bool} We consider all the cases next:

\begin{itemize}
\item Case $\varphi \wedge \psi$.
  We remind the reader that $\varphi \wedge \psi \eqdef \deventually{\varphi?}{\psi}$. The proof goes as follows:
	\begin{align*}
		& \phantom{\text{iff}} \M, k \models \deventually{\varphi?}{\psi} \\
		& \text{ iff }  \M, k \models \mdeventually{\varphi?}{\intervoo{-\omega}{\omega}}{\psi}                                                           & \text{by definition of } \deventually{\rho}{\varphi} \text{ in } \eqref{def:dht:mdht} \\
		& \text{ iff } \M,i \models \psi \text{ for some } i \text{ with } (k,i) \in \mdrel{\varphi?}{\M} \\ &
		 \quad\text{ and } \tmf(i)-\tmf(k) \in \intervoo{-\omega}{\omega}  & \text{by Definition}~\ref{def:mdht:satisfaction}\eqref{def:mdhtsat:eventually:f}\\
		& \text{ iff } \M,i \models \psi \text{ for some } i \text{ with } (k,i) \in \mdrel{\varphi?}{\M} & \text{ since } \tmf(i)-\tmf(k) \in \mathbb{Z}\\
		& \text{ iff } \M,k \models \varphi \text{ and } \M,k \models \psi  & \text{ since } (k,k) \in \mdrel{\varphi?}{\M}	\\
		& \text{ iff } \M,k \models \varphi\wedge \psi. & 	\\
	\end{align*}
      \item Case $\varphi \vee \psi$.
        We remind the reader that $\varphi \vee \psi \eqdef \deventually{\varphi?+\psi?} \top$. The proof is presented below.
  \begin{align*}
	    &\phantom{\text{iff}} \M, k \models \deventually{\varphi?+\psi?} \top   \\
	    & \text{ iff }  \M, k \models \mdeventually{\varphi?+\psi?}{\intervoo{-\omega}{\omega}}{\top}                                                                         & \text{by definition of } \deventually{\rho}{\varphi} \text{ in } \eqref{def:dht:mdht}\\
		& \text{ iff } \M,i \models \top \text{ for some } (k,i) \in \mdrel{\varphi?+\psi?}{\M}& \\ & \quad \text{ and } \tmf(i)-\tmf(k) \in \intervoo{-\omega}{\omega} & \text{by Definition}~\ref{def:mdht:satisfaction}\eqref{def:mdhtsat:eventually:f}\\
		& \text{ iff } \text{ there exists } (k,i) \in \mdrel{\varphi?+\psi?}{\M}                                                              & \text{ since }\tmf(i)-\tmf(k) \in \mathbb{Z}\\
		& \text{ iff there exists } (k,i) \in \mdrel{\varphi?}{\M}\cup\mdrel{\psi?}{\M} & \mdrel{\varphi?+\psi?}{\M} \eqdef \mdrel{\varphi?}{\M}\cup\mdrel{\psi?}{\M} \\
		& \text{ iff } k=i \text{ and } \M, k \models \varphi \hbox{ or } \M, k \models \psi & \text{ definition of } \mdrel{\varphi?}{\M} \text{ and } \mdrel{\psi?}{\M}\\
		& \M, k \models \varphi \vee \psi
 \end{align*}
\item Case $\varphi \to \psi$.
  We remind the reader that $\varphi \to \psi \eqdef  \dalways{\varphi?} \psi$ and we present the proof next
 \begin{align*}
	 	     & \phantom{\text{iff}} \M, k \models \dalways{\varphi?} \psi    \\
		     & \text{ iff } \M, k \models \mdalways{\varphi?}{\intervoo{-\omega}{\omega}}{\psi}                                & \text{by definition of } \dalways{\rho}{\varphi} \text{ in } \eqref{def:dht:mdht}\\
	 	     & \text{ iff } \M',i \models \psi \text{ for all } (k,i) \in \mdrel{\varphi?}{\M'} \text{ satisfying }                                                & \\
	 	     & \quad   \tmf(i)-\tmf(k) \in \intervoo{-\omega}{\omega} \text{ and for both }& \\
	 	     & \quad\M'=\M \text{ and } \M'=\tuple{\Ttrace,\Ttrace,\tau} & \text{by Definition}~\ref{def:mdht:satisfaction}\eqref{def:mdhtsat:always:f}\\
	 	     & \text{ iff for all } i \text{ we have } \M',i \models \psi  \text{ if } i=k \text{ and } \M' \models \varphi & \text{ since } (k,i) \in \mdrel{\varphi?}{\M'} \\
	 	     & \quad \text{for both }\M'=\M \text{ and } \M'=\tuple{\Ttrace,\Ttrace,\tmf}&  \text{ and }  \tmf(i)-\tmf(k) \in \mathbb{Z}\\
	 	     &\text{ iff } \M',k \models \psi  \text{ if }  \M' \models \varphi \text{ for both } & \\
			 & \quad \M'=\M \text{ and } \M'=\tuple{\Ttrace,\Ttrace,\tmf} & \text{ since } i=k\\
	 	     & \text{ iff }\M, k \models \varphi \rightarrow \psi.
	  \end{align*}
\end{itemize}

\end{proof}
%

%% file: proofs/prop-persistence.tex
\begin{proof}\Proposition{prop:persistence}

\noindent	We proceed by induction on the well-founded subexpression relation taking into account the following considerations: for any arbitrary subexpression $\gamma$ of an expression $\varphi$:

	\begin{itemize}
		\item[(A)] Persistence holds if $\gamma$ is a formula and 
		\item[(B)] $\mdrel{\gamma}{\tuple{{\Htrace,\Ttrace},\tau}} \subseteq \mdrel{\gamma}{\tuple{{\Ttrace,\Ttrace},\tau}}$ holds if $\gamma$ is a path expression.
	\end{itemize}

  First we show (A) by assuming that the induction hypothesis (A) and (B) hold for proper subexpressions of $\varphi$.

 \begin{itemize}
 	\item Case of $\varphi= a \in \PV$: if $\tuple{\Htrace,\Ttrace,\tau},k \models a$ then $a \in \Htrace_k$ by the \HT{} semantics. Since $\Htrace_k \subseteq  \Ttrace_k$ then $a \in \Ttrace_k$. 
 	Finally, $\tuple{\Ttrace,\Ttrace,\tau},k \models a$ because of the \HT{} semantics.
 	\item Case of $\varphi = \bot$: it trivially holds that $\tuple{\Ttrace,\Ttrace, \tau}, k \not \models \bot$.
 	\item Case of $\varphi = \mdeventually{\rho}{I}\varphi$: if $\tuple{\Htrace,\Ttrace,\tau},k \models \mdeventually{\rho}{I}\varphi$ then  $\tuple{\Htrace,\Ttrace,\tmf},i \models \varphi$ for some $(k,i) \in \mdrel{\rho}{\tuple{\Htrace,\Ttrace,\tau}}$ and $\tmf(i)-\tmf(k) \in I$, because of the \HT{} semantics.
 	
 	By applying the induction hypothesis (A) on $\varphi$ and (B) on $\rho$ we get that $\tuple{\Ttrace,\Ttrace,\tmf},i \models \varphi$ and $(k,i) \in \mdrel{\rho}{\tuple{\Ttrace,\Ttrace,\tau}}$.
 	Finally, by the \HT{} semantics we get $\tuple{\Ttrace, \Ttrace,\tmf}, k \models \mdeventually{\rho}{I}\varphi$.
 
 	\item Case of $\varphi = \mdalways{\rho}{I}\varphi$: if $\tuple{\Htrace,\Ttrace,\tmf},k \models \mdalways{\rho}{I}\varphi$ then  $\tuple{\Htrace,\Ttrace,\tmf},i \models \varphi$
 	for all $(k,i) \in \mdrel{\rho}{\M'}$ satisfying $\tmf(i)-\tmf(k) \in I$ and for both $\M'=\tuple{\Htrace,\Ttrace,\tmf}$ and $\M'=\tuple{\Ttrace,\Ttrace,\tmf}$. By the \HT{} semantics we directly get $\tuple{\Ttrace,\Ttrace,\tmf}, k \models \mdalways{\rho}{I}\varphi$\footnote{Notice that the \HT{} semantics forces formulas of the type $\mdalways{\rho}{I}\varphi$ to be satisfied in the ``there'' component. Therefore no induction hypothesis is needed in this case.}
 \end{itemize}

To prove the property (B) we prove the following equivalent property: for all $(k,i) \in \mathbb{N} \times \mathbb{N}$ and for all path expressions $\rho$

\begin{displaymath}
	(k,i) \in \mdrel{\rho}{\tuple{\Htrace,\Ttrace,\tmf}} \hbox{ implies } (k,i) \in \mdrel{\rho}{\tuple{\Ttrace,\Ttrace,\tmf}}.
\end{displaymath}

	\begin{itemize}
		\item Case $\rho = \stp$: if $(k,i) \in \mdrel{\stp}{\tuple{\Htrace,\Ttrace,\tmf}}$ then, by definition, $i = k+1$. Again, by the definition of $\mdrel{\stp}{\tuple{\Ttrace,\Ttrace,\tmf}}$, $(k,i) \in \mdrel{\stp}{\tuple{\Ttrace,\Ttrace,\tmf}}$. 
		\item Case $\rho = \varphi ?$: if $(k,i) \in \mdrel{\varphi?}{\tuple{\Htrace,\Ttrace,\tmf}}$ then, by definition $i=k$ and $\tuple{\Htrace,\Ttrace,\tmf}, k \models \varphi$. By applying the induction hypothesis (A) on $\varphi$ we get $\tuple{\Ttrace,\Ttrace,\tmf}, k \models \varphi$. Finally, by the definition of $\mdrel{\varphi?}{\tuple{\Htrace,\Ttrace,\tmf}}$, $(k,i) \in \mdrel{\varphi?}{\tuple{\Ttrace,\Ttrace,\tmf}}$. 
		\item Case $\rho = \rho_1 + \rho_2$: if $ (k,i) \in \mdrel{\rho_1\mathrel{+}\rho_2}{\tuple{\Htrace,\Ttrace,\tmf}}$ then, by the definition of $\mdrel{\rho_1\mathrel{+}\rho_2}{\tuple{\Htrace,\Ttrace,\tmf}}$, either $ (k,i) \in \mdrel{\rho_1}{\tuple{\Htrace,\Ttrace,\tmf}}$ or $ (k,i) \in \mdrel{\rho_2}{\tuple{\Htrace,\Ttrace,\tmf}}$. By applying the induction hypothesis (B) on $\rho_1$ and $\rho_2$ we obtain that either $(k,i) \in \mdrel{\rho_1}{\tuple{\Ttrace,\Ttrace,\tmf}}$ or $ (k,i) \in \mdrel{\rho_2}{\tuple{\Ttrace,\Ttrace,\tmf}}$. By the definition of $\mdrel{\rho_1\mathrel{+}\rho_2}{\tuple{\Ttrace,\Ttrace,\tmf}}$ it follows that $(k,i) \in \mdrel{\rho_1\mathrel{+}\rho_2}{\tuple{\Htrace,\Ttrace,\tmf}}$
		\item Case $\rho = \rho_1 ; \rho_2$: if $(k,i) \in \mdrel{\rho_1;\rho_2}{\tuple{\Htrace,\Ttrace,\tmf}}$ then $(k,j) \in \mdrel{\rho_1}{\tuple{\Htrace,\Ttrace,\tmf}}$ and $(j,i) \in \mdrel{\rho_2}{\tuple{\Htrace,\Ttrace,\tmf}}$ for some $i \in \mathbb{N}$. If we apply the induction hypothesis (B) on $\rho_1$ and $\rho_2$ we get $(k,j) \in \mdrel{\rho_1}{\tuple{\Ttrace,\Ttrace,\tmf}}$ and $(j,i) \in \mdrel{\rho_2}{\tuple{\Ttrace,\Ttrace,\tmf}}$. By the definition of $\mdrel{\rho_1;\rho_2}{\tuple{\Htrace,\Ttrace,\tmf}}$ we get $(k,i) \in \mdrel{\rho_1;\rho_2}{\tuple{\Ttrace,\Ttrace,\tmf}}$
		\item Case $\rho=\rho^-$: if $(k,i) \in \mdrel{\converse{\rho}}{\tuple{\Htrace,\Ttrace,\tmf}}$ then $(i,k) \in \mdrel{\rho}{\tuple{\Htrace,\Ttrace,\tmf}}$ by definition. By applying the induction hypothesis (B) on $\rho$ we get that $(i,k) \in \mdrel{\rho}{\tuple{\Ttrace,\Ttrace,\tmf}}$ and, again, by definition $(k,i) \in \mdrel{\converse{\rho}}{\tuple{\Ttrace,\Ttrace,\tmf}}$
		\item Case $\rho= \rho^*$: Let us claim that :  
		\begin{equation}
		\text{for all }	n \in \mathbb{N},\; \mdrel{\rho^n}{\tuple{\Htrace,\Ttrace,\tmf}} \subseteq \mdrel{\rho^n}{\tuple{\Ttrace,\Ttrace,\tmf}}\label{eq:ind:n}
		\end{equation}
	
		\noindent holds. If $(k,i) \in \mdrel{\rho^*}{\tuple{\Htrace,\Ttrace,\tmf}}$ then $(k,i) \in \mdrel{\rho^n}{\tuple{\Htrace,\Ttrace,\tmf}}$ for some $n\in \mathbb{N}$. By~\eqref{eq:ind:n}, $(k,i) \in \mdrel{\rho^n}{\tuple{\Ttrace,\Ttrace,\tmf}}$.By the definition of $\mdrel{\rho^*}{\tuple{\Ttrace,\Ttrace,\tmf}}$, $(k,i)\in \mdrel{\rho^*}{\tuple{\Ttrace,\Ttrace,\tmf}}$.
		
		To finish the proof of the property (B) we need to prove~\eqref{eq:ind:n}, which is done by induction on $\mathbb{N}$.
		
		\begin{itemize}
			\item Case $n = 0$: if $(k,i) \in \mdrel{\rho^0}{\tuple{\Htrace,\Ttrace,\tmf}}$ then $(k,i) \in \mdrel{\top?}{\tuple{\Htrace,\Ttrace,\tmf}}$. By definition, $i=k$. By the definition of $ \mdrel{\rho^0}{\tuple{\Ttrace,\Ttrace,\tmf}}$ it follows that $(k,i) \in \mdrel{\rho^0}{\tuple{\Ttrace,\Ttrace,\tmf}}$
			\item inductive step: Assume that~\eqref{eq:ind:n} holds for all $0\le i \le n$ and let us prove it for $n+1$. If $(k,i) \in \mdrel{\rho^{n+1}}{\tuple{\Htrace,\Ttrace,\tmf}}$ then $(k,i) \in \mdrel{\rho;\rho^n}{\tuple{\Htrace,\Ttrace,\tmf}}$ by definition. Consequently, $(k,j) \in \mdrel{\rho}{\tuple{\Htrace,\Ttrace,\tmf}}$ and  $(j,i) \in \mdrel{\rho^n}{\tuple{\Htrace,\Ttrace,\tmf}}$ for some $j\in \mathbb{N}$. By applying the induction hypothesis (B) on $\rho$ and the induction on $n$ we conclude that $(k,j) \in \mdrel{\rho}{\tuple{\Ttrace,\Ttrace,\tmf}}$ and  $(j,i) \in \mdrel{\rho^n}{\tuple{\Ttrace,\Ttrace,\tmf}}$. By definition of $\mdrel{\rho^{n+1}}{\tuple{\Ttrace,\Ttrace,\tmf}}$ we get that $ (k,i) \in \mdrel{\rho^{n+1}}{\tuple{\Ttrace,\Ttrace,\tmf}}$.
		\end{itemize}
	\end{itemize}

\end{proof}

%% file: proofs/prop-totality.tex
\begin{proof}\Proposition{prop:totality}
Proof by induction on the well-founded subexpression relation.
As a remark, note that under the assumption of total traces, the law of excluded middle is satisfied
and this makes that \MDHT{} formulas to be satisfied as in \MDL{}.
\end{proof}
%

%% file: proofs/prop-implications.tex
\begin{proof}\Proposition{prop:no:implication}
By induction on the well-founded subexpression relation.
Note that the absence of implications implies that the world `here' is the only one used in the evaluation of a formula.
\end{proof}

%% file: proofs/prop-satisfaction-dht.tex
\begin{proof}\Proposition{prop:satisfaction:dht}
We consider all the cases below:

\begin{itemize}
	\item Case $\deventually{\rho}{\varphi}$: we remind the reader that $\deventually{\rho}{\varphi}\eqdef \mdeventually{\rho}{\intervoo{-\omega}{\omega}}{\varphi}$. The proof is presented next.
	
	  \begin{align*}
		\M, k \models   \mdeventually{\rho}{\intervoo{-\omega}{\omega}}{\varphi}                        & \text{ iff }  \M,i \models \varphi
		\text{ for some } (k,i) \in \mdrel{\rho}{\M} & \\
		& \quad \text{ such that } \tmf(i)-\tmf(k) \in \intervoo{-\omega}{\omega}
		& \text{by Definition}~\ref{def:mdht:satisfaction}\eqref{def:mdhtsat:eventually:f}\\
		& \text{ iff } \M,i \models \varphi
		\text{ for some } (k,i) \in \mdrel{\rho}{\M}                                                                   & \text{since }\tmf(i)-\tmf(k)\in\mathbb{Z}\\
	\end{align*}
	
	\item Case $\dalways{\rho}{\varphi}$: we remind the reader that $\dalways{\rho}{\varphi}\eqdef \mdalways{\rho}{\intervoo{-\omega}{\omega}}{\varphi}$. The proof is presented below.
	
	  \begin{align*}
		\mdalways{\rho}{\intervoo{-\omega}{\omega}}{\varphi} & \text{ iff } \M',i \models \varphi
		\text{ for all } (k,i) \in \mdrel{\rho}{\M'}                                                                    & \\
		& \quad		
		\text{satisfying }
		\tmf(i)-\tmf(k) \in \intervoo{-\omega}{\omega}                                                      & \\
		& \quad
		\text{and for both } \M'=\tuple{\Htrace,\Ttrace,\tmf} \text{ and } \M'=\tuple{\Ttrace,\Ttrace,\tmf} & \text{by Definition}~\ref{def:mdht:satisfaction}\eqref{def:mdhtsat:always:f}\\
		& \text{ iff } \M',i \models \varphi
		\text{ for all } (k,i) \in \mdrel{\rho}{\M'}                                                                         & \\
		& \quad
		\text{and for both } \M'=\tuple{\Htrace,\Ttrace,\tmf} \text{ and } \M'=\tuple{\Ttrace,\Ttrace,\tmf} & \text{since }\tmf(i)-\tmf(k)\in\mathbb{Z}\\
	\end{align*}

\end{itemize}

\end{proof}
%

%% file: proofs/prop-satisfaction-mht.tex
\begin{proof}\Proposition{prop:satisfaction:mht}
Before presenting the proof, let us introduce the following considerations: 

\begin{enumerate}
	\item We will work with a timed trace  $\M = $ of length $\lambda$;
	\item We will denote by $\mathbb{D} \eqdef \lbrace i \mid 0 \le i < \lambda \rbrace$ the possible states of $\M$ and
	\item We will implicitly consider that $m,n\in \mathbb{N}$.
\end{enumerate}

\noindent Notice that, for any path expression $\rho$ and for any model $\M$, $\mdrel{\rho}{\M} \subseteq \mathbb{D}\times\mathbb{D}$. 
We consider all cases next:

\begin{itemize}
	\item Case $\initially$: note that $\initially \eqdef \dalways{ \converse{\stp}}{\bot}$ by Definition~\eqref{def:mht:final:initial}. Therefore $\initially \eqdef \mdalways{ \converse{\stp}      }{\intervoo{-\omega}{\omega}}{\bot} $ because of Definition~\eqref{def:mdhtsat:always:f}. The rest of the proof goes as follows:

  \begin{align*}	
	& \M,k \models \mdalways{ \converse{\stp}}{\intervoo{-\omega}{\omega}}{\bot} \text{ iff } \M',i \models \bot \text{ if }	(k,i) \in \mdrel{\converse{\stp}}{\M'}  \text{ and } &\\
	& \quad \tmf(i)-\tmf(k) \in \intervoo{-\omega}{\omega},\; \text{for all } (k,i) \in \mathbb{D} \times \mathbb{D}  &  \\
	& \quad \text{ and } \M' \in \lbrace\M, \tuple{\Ttrace,\Ttrace,\tmf} \rbrace &\text{by the satisfaction of } \metricI{\dalways{\rho}}\varphi \\
	& \text{ iff } \M',i \models \bot \text{ if }	(i,k) \in \mdrel{\stp}{\M'} \text{ and } \tmf(i)-\tmf(k) \in \intervoo{-\omega}{\omega}, &\\
	& \quad  \text{for all } (k,i) \in \mathbb{D} \times \mathbb{D} \text{ and } \M' \in \lbrace \M, \tuple{\Ttrace,\Ttrace,\tmf} \rbrace & \text{by the definition of }\mdrel{\stp^-}{\M'}\\
	& \text{ iff } \M',i \models \bot \text{ if }	i = k-1 \text{ and } \tmf(i)-\tmf(k) \in \intervoo{-\omega}{\omega}, &\\
	& \quad  \text{for all } (k,i) \in \mathbb{D} \times \mathbb{D} \text{ and }\M' \in \lbrace \M, \tuple{\Ttrace,\Ttrace,\tmf} \rbrace & \text{by the definition of }\mdrel{\stp}{\M'}\\
	& \text{ iff } \M',i \models \bot \text{ if }	i = k-1 \text{ for all } (k,i) \in \mathbb{D} \times \mathbb{D}&\\
	& \quad \text{ and }\M' \in \lbrace \M, \tuple{\Ttrace,\Ttrace,\tmf} \rbrace & \tmf(i)-\tmf(k) \in \intervoo{-\omega}{\omega}\\
	& \text{ iff } k = 0. 
\end{align*}

	\item Case of $\finally$: note that $\finally \eqdef \dalways{ \stp}{\bot}$ by Definition of $\finally$. Since $\dalways{ \stp}{\bot} \equiv \mdalways{ \stp     }{\intervoo{-\omega}{\omega}}{\bot}$, we conclude that $\finally \equiv \mdalways{ \stp     }{\intervoo{-\omega}{\omega}}{\bot}$. The rest of the proof goes as follows.

  \begin{align*}	
	& \M,k \models \mdalways{\stp}{\intervoo{-\omega}{\omega}}{\bot} \text{ iff } \M',i \models \bot \text{ if }	(k,i) \in \mdrel{\stp}{\M'} \text{ and }  &\\
	& \quad \tmf(i)-\tmf(k) \in \intervoo{-\omega}{\omega},\; \text{for all } (k,i) \in \mathbb{D} \times \mathbb{D} & \\
	& \quad \text{ and } \M' \in \lbrace \M, \tuple{\Ttrace,\Ttrace,\tmf} \rbrace & \text{by the satisfaction of }\metricI{\dalways{\rho}}\varphi \\
	& \text{ iff } \M',i \models \bot \text{ if }	i = k+1 \text{ and } \tmf(i)-\tmf(k) \in \intervoo{-\omega}{\omega} &\\
	& \quad  \text{for all } (k,i) \in \mathbb{D} \times \mathbb{D}  \text{ and } \M' \in \lbrace \M, \tuple{\Ttrace,\Ttrace,\tmf} \rbrace & \text{by definition of }\mdrel{\stp}{\M'}\\
	& \text{ iff } \M',i \models \bot \text{ if }	i = k+1,\; \text{ for all } (k,i) \in \mathbb{D} \times \mathbb{D} &\\
	& \quad \text{ and } \M' \in \lbrace \M, \tuple{\Ttrace,\Ttrace,\tmf} \rbrace & \tmf(i)-\tmf(k) \in \intervoo{-\omega}{\omega}\\
	& \text{ iff } k = \lambda-1.
\end{align*}

\item Case of $\next_{[m,n)} \varphi$: note that $\next_{[m,n)} \varphi \eqdef \deventually{\stp}_{[m,n)}\varphi$. Having this in mind, we present the proof below

\begin{align*}
	\M,k \models  \deventually{\stp}_{[m,n)}\varphi & \text{ iff } \M,i \models \varphi
	\text{ for some } (k,i) \in \mdrel{\stp}{\M} \\
	&\quad \text{ and } \tmf(i)-\tmf(k) \in [m,n)&\text{by the satisfaction of} \metricI{\deventually{\rho}}\varphi\\
	& \text{ iff } \M,i \models \varphi
	\text{ and } 0\le i=k+1 < \lambda \text{ and } \\
	&\quad \tmf(i)-\tmf(k) \in [m,n)&\text{by the definition of } \mdrel{\stp}{\M} \\	
	& \text{ iff } k+1 < \lambda \text{ and } \M,k+1 \models \varphi \\
	& \quad \text{ and } \tmf(k+1)-\tmf(k) \in [m,n)& i=k+1 < \lambda\\		
	& \text{ iff } \M, k \models \next_{[m,n)} \varphi.
\end{align*}

\item Case of $\wnext_{[m,n)} \varphi$: note that $\wnext_{[m,n)} \varphi \eqdef \dalways{\stp}_{[m,n)}\varphi$. Having this in mind, we present the proof below

\begin{align*}
	\M,k \models  \dalways{\stp}_{[m,n)}\varphi & \text{ iff } \M',i \models \varphi
	\text{ if } (k,i) \in \mdrel{\stp}{\M'}  \text{ and } & \\
	&\quad \tmf(i)-\tmf(k) \in [m,n) \text{ for all }  (k,i) \in \mathbb{D}\times\mathbb{D} & \\
	&\quad  \text{ and } \M'\in \lbrace\M,\tuple{\Ttrace,\Ttrace,\tmf} \rbrace &  \text{satisfaction of } \metricI{\dalways{\rho}}\varphi\\ 
	& \text{ iff } \M',i \models \varphi \text{ if } 0 \le i = k+1 < \lambda  \text{ and } &  \\
	&\quad  \tmf(i)-\tmf(k) \in [m,n), \text{ for  all } \M' \in \lbrace\M,\tuple{\Ttrace,\Ttrace,\tmf} \rbrace & \text{definition of } \mdrel{\stp}{\M'} \\ 
	& \text{ iff } \M',k+1 \models \varphi \text{ if } k+1 < \lambda  \text{ and }   \\
	&\quad  \tmf(k+1)-\tmf(k) \in [m,n) \text{ for  all } \M' \in \lbrace\M,\tuple{\Ttrace,\Ttrace,\tmf} \rbrace & i = k+1 \\ 
	& \text{ iff } \M, k \models \wnext_{[m,n)} \varphi
\end{align*}

\item Case of $\previous_{[m,n)} \varphi$: note that $\previous_{[m,n)} \varphi \eqdef \deventually{\converse{\stp}}_{(-n,-m]}\varphi$. Having this in mind, we present the proof below

  \begin{align*}
	\M,k \models  \deventually{\converse{\stp}}_{(-n,-m]}\varphi & \text{ iff } \M,i \models \varphi
	\text{ for some } (k,i) \in \mdrel{\converse{\stp}}{\M} \\
	&\quad \text{ and } \tmf(i)-\tmf(k) \in (-n,-m]&\text{by the satisfaction of} \metricI{\deventually{\rho}}\varphi \\
	& \text{ iff } \M,i \models \varphi
	\text{ for some } (i,k) \in \mdrel{\stp}{\M} \\
	&\quad \text{ and } \tmf(i)-\tmf(k) \in (-n,-m]&\text{by the definition of } \mdrel{\converse{\stp}}{\M}  \\
	& \text{ iff } \M,i \models \varphi
	\text{ and } \lambda > i=k-1 \ge 0 \text{ and } \\
	&\quad \tmf(i)-\tmf(k) \in (-n,-m]&\text{by Definition of } \mdrel{\stp}{\M} \\	
	& \text{ iff } \M,i \models \varphi
	\text{ and } \lambda > i=k-1 \ge 0  \text{ and } \\
	&\quad \tmf(k)-\tmf(i) \in [m,n)&\text{by Remark~\ref{remark:inecuations}}\\	
	& \text{ iff } \lambda > k-1 \ge 0 \text{ and } \M,k-1 \models \varphi \\
	& \quad \text{ and } \tmf(k)-\tmf(k-1) \in [m,n)& i=k-1 > 0\\		
	& \text{ iff } \M, k \models \previous_{[m,n)} \varphi.
\end{align*}
	
\item Case of $\wprevious_{[m,n)} \varphi$: note that $\wprevious_{[m,n)} \varphi \eqdef \dalways{\converse{\stp}}_{(-n,-m]}\varphi$. Having this in mind, we present the proof below

\begin{align*}
	\M,k \models  \dalways{\converse{\stp}}_{(-n,-m]}\varphi & \text{ iff } \M',i \models \varphi
	\text{ if } (k,i) \in \mdrel{\converse{\stp}}{\M'} \\
	& \quad \text{ and } \tmf(i)-\tmf(k) \in (-n,-m],\\
	&\quad \text{ for all } (k,i) \in \mathbb{D} \times \mathbb{D} \text{ and } \M' \in \lbrace \M,\tuple{\Ttrace,\Ttrace,\tau}\rbrace  & \text{satisfaction of } \metricI{\dalways{\rho}} \varphi\\
	& \text{ iff } \M',i \models \varphi\text{ if } (i,k) \in \mdrel{\stp}{\M'}\\
	&\quad  \text{ and } \tmf(i)-\tmf(k) \in (-n,-m],\\
	&\quad \text{ for all } (k,i) \in \mathbb{D} \times \mathbb{D} \text{ and } \M' \in \lbrace \M,\tuple{\Ttrace,\Ttrace,\tau}\rbrace  & \text{definition of } \drel{\converse{\rho}}{\M'}\\
	& \text{ iff } \M',i \models \varphi
	\text{ if } \lambda > i = k-1 \ge 0  \text{ and }\\
	&\quad \tmf(i)-\tmf(k) \in (-n,-m], \\
	& \quad \text{ for all } \M' \in \lbrace \M,\tuple{\Ttrace,\Ttrace,\tau}\rbrace  & \text{definition of } \drel{\rho}{\M'}\\
	& \text{ iff } \M',i \models \varphi
	\text{ if } \lambda > i = k-1 \ge 0  \text{ and }\\
	&\quad  \tmf(k) - \tmf(i) \in [m,n) \text{, for all } \M' \in \lbrace \M,\tuple{\Ttrace,\Ttrace,\tau}\rbrace  & \text{by Remark~\ref{remark:inecuations}}\\
	& \text{ iff } \M',k-1 \models \varphi
	\text{ if } k-1 \ge 0  \text{ and } \\ 
	&\quad \tmf(k) - \tmf(k-1) \in [m,n), \\
	& \quad \text{ for all } \M' \in \lbrace \M,\tuple{\Ttrace,\Ttrace,\tau}\rbrace  & i = k-1\\
\end{align*}

\item Case of $\eventuallyF_{[m,n)} \varphi$: as a reminder, $\eventuallyF_{[m,n)} \varphi \eqdef \deventually{\stp^*}_{[m,n)} \varphi$ as follows
		
	\begin{align*}
			    \M,k \models \deventually{\stp^*}_{[m,n)}\varphi & \text{ iff } \M,i \models \varphi \text{ for some } (k,i) \in \drel{\stp^*}{\M}\\
			    												 & \quad \text{ with } \tmf(i) - \tmf(k) \in [m,n) &  \text{satisfaction of } \metricI{\deventually{\rho}}\varphi\\
																 & \text{ iff } \M,i \models \varphi \text{ for some } (k,i) \in \drel{\stp^x}{\M}\\
																 & \quad \text{ with } x \ge 0 \text{ and } \tmf(i) - \tmf(k) \in [m,n) &  \text{definition of } \drel{\rho^*}{\M}\\
																 & \text{ iff } \M,i \models \varphi \text{ for some } 0 \le i = k+x < \lambda\\
																 & \quad \text{ with } x \ge 0 \text{ and }  \tmf(i) - \tmf(k) \in [m,n) &  \text{by Proposition~\ref{prop:aux:ref}}\\																 
																 & \text{ iff } \M,i \models \varphi \text{ for some }  i- k \ge 0 \text{ satisfying } 0 \le i < \lambda\\
																 & \quad  \text{ and } \tmf(i) - \tmf(k) \in [m,n) & i = k+x\\								
																 & \text{ iff } \M,i \models \varphi \text{ for some }  0 \le k \le i < \lambda  \\																	 
																 & \quad \text{ and } \tmf(i) - \tmf(k) \in [m,n) & i-k \ge 0\\
																 & \text{ iff } \M,k \models  \eventuallyF_{[m,n)} \varphi. \\																																 																															 
\end{align*}

\item Case of $\alwaysF_{[m,n)} \varphi$: as a reminder, $\alwaysF_{[m,n)} \varphi \eqdef \dalways{\stp^*}_{[m,n)} \varphi$ as follows

\begin{align*}
	\M,k \models \dalways{\stp^*}_{[m,n)}\varphi & \text{ iff } \M',i \models \varphi  \text{ if } (k,i) \in \drel{\stp^*}{\M'} \\
	& \quad \text{ and  } \tmf(i) - \tmf(k) \in [m,n) \text{, for all }  & \\ 
	& \quad (k,i) \in \mathbb{D} \times \mathbb{D} \text{ and } \M' \in \lbrace \M, \tuple{\Ttrace,\Ttrace,\tmf} \rbrace&  \text{satisfaction of } \metricI{\dalways{\rho}}\varphi\\
	& \text{ iff } \M',i \models \varphi  \text{ if } x \ge 0 \text{, }(k,i) \in \drel{\stp^x}{\M'}\\
	& \quad \text{ and }  \tmf(i) - \tmf(k) \in [m,n) \text{, for all } x\in \mathbb{N}, &  \\ 
	& \quad (k,i) \in \mathbb{D} \times \mathbb{D} \text{ and } \M' \in \lbrace \M, \tuple{\Ttrace,\Ttrace,\tmf} \rbrace &   \text{definition of } \drel{\rho^*}{\M'}\\
	& \text{ iff } \M',i \models \varphi  \text{ if } x \ge 0,\ 0 \le  i= k+x < \lambda \\
	& \quad \text{ and }  \tmf(i) - \tmf(k) \in [m,n) \text{, for all } x \in \mathbb{N} &  \\ 
	& \quad \text{ and } \M' \in \lbrace \M, \tuple{\Ttrace,\Ttrace,\tmf} \rbrace &    \text{By Proposition~\ref{prop:aux:ref}}\\	
	& \text{ iff } \M',i \models \varphi  \text{ if } i-k \ge 0,\ 0 \le i < \lambda \text{ and } \\
	& \quad   \tmf(i) - \tmf(k) \in [m,n), \text{ for all } \M' \in \lbrace \M, \tuple{\Ttrace,\Ttrace,\tmf} \rbrace &    i= k+x \\		
	& \text{ iff } \M',i \models \varphi  \text{ if } k \le i < \lambda \text{ and }  \tmf(i) - \tmf(k) \in [m,n), \\
	& \quad \hbox{ for all } \M' \in \lbrace \M, \tuple{\Ttrace,\Ttrace,\tmf} \rbrace &    i-k \ge 0 \\	
	& \text{ iff } \M, k \models \alwaysF_{[m,n)} \varphi
\end{align*}

\item Case of $\eventuallyP_{[m,n)} \varphi$: as a reminder, $\eventuallyP_{[m,n)} \varphi \eqdef \deventually{(\stp^*)^-}_{(-n,-m]}$ as follows

\begin{align*}
		\M,k \models \deventually{(\stp^*)^-}_{[m,n)}\varphi & \text{ iff } \M,i \models \varphi \text{ for some } (k,i) \in \drel{(\stp^*)^-}{\M}\\
			& \quad \text{ with } \tmf(k) - \tmf(i) \in (-n,-m] &  \text{satisfaction of } \metricI{\deventually{\rho}}\varphi\\
			& \text{ iff } \M,i \models \varphi \text{ for some } (i,k) \in \drel{\stp^*}{\M}\\
			& \quad \text{ with } \tmf(i) - \tmf(k) \in (-n,-m] &  \text{definition of } \drel{\rho^-}{\M}\\						
			& \text{ iff } \M,i \models \varphi \text{ for some } (i,k) \in \drel{\stp^x}{\M} & \\
			& \quad \text{ with } x \ge 0 \text{ and } \tmf(i) - \tmf(k) \in (-n,-m] &  \text{definition of } \drel{\rho^*}{\M}\\
			& \text{ iff } \M,i \models \varphi \text{ for some } 0\le i = k-x < \lambda \\
			& \quad \text{ with } x \ge 0 \text{ and } \tmf(i) - \tmf(k) \in (-n,-m] &  \text{by Proposition~\ref{prop:aux:ref}}	\\																 
			& \text{ iff } \M,i \models \varphi \text{ for some } 0\le i < \lambda\\
			& \quad \text{ with } k-i \ge 0 \text{ and } \tmf(i) - \tmf(k) \in (-n,-m] &  i = k-i	\\																 			
			& \text{ iff } \M,i \models \varphi \text{ for some } 0\le i \le k < \lambda\\
			& \quad \text{ with } \tmf(i) - \tmf(k) \in (-n,-m] &  \text{since  }   k-i \ge 0	\\																 									
			& \text{ iff } \M,i \models \varphi \text{ for some } 0\le i \le k < \lambda\\
			& \quad \text{ with } \tmf(k) - \tmf(i) \in [m,n) &  \text{because of  Remark~\ref{remark:inecuations}}  	\\
			& \text{ iff } \M,k \models  \eventuallyP_{[m,n)} \varphi\\								
\end{align*}

\item Case of $\alwaysP_{[m,n)} \varphi$: as a reminder, $\alwaysP_{[m,n)} \varphi \eqdef \dalways{(\stp^*)^-}_{(-n,-m]} \varphi$ as follows

\begin{align*}
	\M,k \models \dalways{(\stp^*)^-}_{(-n,-m]}\varphi & \text{ iff } \M',i \models \varphi  \text{ if } (k,i) \in \drel{(\stp^*)^-}{\M'} \\
	& \quad \text{ and  } \tmf(i) - \tmf(k) \in (-n,-m] \text{, for all }  & \\ 
	& \quad (k,i) \in \mathbb{D} \times \mathbb{D} \text{ and } \M' \in \lbrace \M, \tuple{\Ttrace,\Ttrace,\tmf} \rbrace&  \text{satisfaction of } \metricI{\dalways{\rho}}\varphi\\
    & \text{ iff } \M',i \models \varphi  \text{ if both } (i,k) \in \drel{\stp^*}{\M'} \\
    & \quad \text{ and  } \tmf(i) - \tmf(k) \in (-n,-m] \text{, for all }  & \\ 
    & \quad (k,i) \in \mathbb{D} \times \mathbb{D} \text{ and } \M' \in \lbrace \M, \tuple{\Ttrace,\Ttrace,\tmf} \rbrace&  \text{satisfaction of } \metricI{\dalways{\rho^-}}\varphi\\
    & \text{ iff } \M',i \models \varphi  \text{ if both } (i,k) \in \drel{\stp^*}{\M'} \\
    & \quad \text{ and  } \tmf(k) - \tmf(i) \in [m,n) \text{, for all }  & \\ 
    & \quad (k,i) \in \mathbb{D} \times \mathbb{D} \text{ and } \M' \in \lbrace \M, \tuple{\Ttrace,\Ttrace,\tmf} \rbrace&  \text{by Remark~\ref{remark:inecuations}}\\        
	& \text{ iff } \M',i \models \varphi  \text{ if } x \ge 0 \text{, }(i,k) \in \drel{\stp^x}{\M'}\\
	& \quad \text{ and }  \tmf(k) - \tmf(i) \in [m,n) \text{, for all } x\in \mathbb{N}, &  \\ 
	& \quad (k,i) \in \mathbb{D} \times \mathbb{D} \text{ and } \M' \in \lbrace \M, \tuple{\Ttrace,\Ttrace,\tmf} \rbrace &   \text{definition of } \drel{\rho^*}{\M'}\\
	& \text{ iff } \M',i \models \varphi  \text{ if } x \ge 0,\ 0 \le  i= k-x < \lambda \\
	& \quad \text{ and }  \tmf(k) - \tmf(i) \in [m,n) \text{, for all } x \in \mathbb{N} &  \\ 
	& \quad \text{ and } \M' \in \lbrace \M, \tuple{\Ttrace,\Ttrace,\tmf} \rbrace &   \text{by Proposition~\ref{prop:aux:ref}}\\	
	& \text{ iff } \M',i \models \varphi  \text{ if } k-i \ge 0,\ 0 \le i < \lambda \\
	& \quad \text{ and }  \tmf(k) - \tmf(i) \in [m,n) \text{, for all } &  \\ 
	& \quad (k,i) \in \mathbb{D} \times \mathbb{D} \text{ and } \M' \in \lbrace \M, \tuple{\Ttrace,\Ttrace,\tmf} \rbrace &    i= k-x \\		
	& \text{ iff } \M',i \models \varphi  \text{ if } 0 \le i \le k < \lambda \text{ and }   \\
	& \quad \tmf(k) - \tmf(i) \in [m,n), \hbox{ for all } (k,i) \in \mathbb{D} \times \mathbb{D} \\	
	& \quad \text{ and } \M' \in \lbrace \M, \tuple{\Ttrace,\Ttrace,\tmf} \rbrace &    k-i \ge 0 \\
	& \text{ iff } \M, k \models \alwaysP_{[m,n)} \varphi
\end{align*}	

\item Case $\varphi \until_{[m,n)} \psi$: we remark that $\varphi \until_{[m,n)} \psi \eqdef \deventually{(\varphi?;\stp)^*}_{[m,n)} \psi$. The proof goes as follows:

\begin{align*}
	\M,k \models \deventually{(\varphi?;\stp)^*}_{[m,n)}\psi & \text{ iff } \M,i \models \psi \text{ for some } (k,i) \in \drel{(\varphi?;\stp)^*}{\M}\\
	& \quad \text{ with } \tmf(i) - \tmf(k) \in [m,n) &  \text{satisfaction of } \metricI{\deventually{\rho}}\varphi\\
	& \text{ iff } \M,i \models \psi \text{ for some } i \ge k \\
	& \quad \text{ s.t. } \tmf(i) - \tmf(k) \in [m,n) & \\
	& \text{ and }\M, j \models \varphi \hbox{ for all } j \in [k,i) & \text{Proposition~\ref{prop:aux:until}}\\
	& \text{ iff } \M,k \models \varphi \until_{[m,n)} \psi & \\
\end{align*}

\item Case $\varphi \since_{[m,n)} \psi$: we remark that $\varphi \since_{[m,n)} \psi \eqdef \deventually{(\varphi?;\stp^-)^*}_{[m,n)} \psi$. The proof goes as follows:

\begin{align*}
	\M,k \models \deventually{(\varphi?;\stp^-)^*}_{[m,n)} \psi & \text{ iff } \M,i \models \psi \text{ for some } (k,i) \in \drel{(\varphi?;\stp^-)^*}{\M}\\
	& \quad \text{ with } \tmf(k) - \tmf(i) \in (-n,-m] &  \text{satisfaction of } \metricI{\deventually{\rho}}\varphi\\
	& \text{ iff } \M,i \models \psi \text{ for some } i \le k \text{ with } \\
	& \quad \tmf(i) - \tmf(k) \in (-n,-m]  & \\
	& \quad \text{ and } \M,j \models \varphi \text{ for all } j \in (i,k] &  \text{by Proposition~\ref{prop:aux:since}}\\
	& \text{ iff } \M,i \models \psi \text{ for some } i \le k \text{ with } \\
	& \quad \tmf(k) - \tmf(i) \in [m,n)  & \\
	& \quad \text{ and } \M,j \models \varphi \text{ for all } j \in (i,k] & \text{by Remark~\ref{remark:inecuations}}\\
\end{align*}					

\item Case $\varphi \release_{[m,n)} \psi$: we remark that in \citep{becadiscsc23} it is shown that
%
$$ \varphi \release_{\intervco{m}{n}} \psi \equiv \alwaysF_{\intervco{m}{n}} \psi \vee \eventuallyF_{\intervco{0}{m}} \left( \varphi \release \left(\varphi \vee \wnext \psi\right)\right).$$ 

In the previous cases we have shown that $\alwaysF_{\intervco{m}{n}}$ and $\eventuallyF_{\intervco{0}{m}}$ are \MDHT-definable.
Furthermore $\release$ is \MDHT-definable by Corollary \ref{cor:dht:tht}.

\item The case $\varphi \trigger_{[m,n)} \psi$ follows a similar reasoning as for $\varphi \release_{[m,n)} \psi$.

\end{itemize}

\end{proof}
%

%% file: proofs/prop-satisfaction-tht.tex
\begin{proof}\Corollary{cor:dht:tht}
The fact that $\THT=\DHT$ for the case of temporal formulas has been proved in~\cite{bocadisc18a}. 
The proof that $\DHT=\MDHT$ for the case of dynamic formulas corresponds to Corollary~\ref{cor:dht:mdht}. 
\end{proof}	
%